\documentclass[10pt,twocolumn,letterpaper]{article}

\usepackage{wacv}
\usepackage{times}
\usepackage{epsfig}
\usepackage{graphicx}
\usepackage{amsmath}
\usepackage{amssymb}

\usepackage{subcaption}
\usepackage{booktabs}	
\usepackage{siunitx}
\usepackage[nolist]{acronym}
\begin{acronym}
    \acro{VAE} {Variational Autoencoder}
    \acro{GAN} {Generative Adversarial Network}
    \acro{MAE} {Mean absolute error}
\end{acronym}

\usepackage{color}
\definecolor{gray_ours}{rgb}{0.6, 0.6, 0.6}
\definecolor{lightblue_ours}{rgb}{0, 0.4470, 0.7410}
\definecolor{greenforest_ours}{rgb}{0.0, 0.9, 0.0}
\definecolor{orangezinnia_ours}{rgb}{1.000, 0.314, 0.020}
\definecolor{yellow_ours}{rgb}{0.9290, 0.6940, 0.1250}

\usepackage{pifont} 	

\usepackage[all]{nowidow}

\newcommand\blfootnote[1]{%
  \begingroup
  \renewcommand\thefootnote{}\footnote{#1}%
  \addtocounter{footnote}{-1}%
  \endgroup
}
\usepackage{textcomp}	
\usepackage[hang,flushmargin]{footmisc}	

\usepackage[pagebackref=true,breaklinks=true,letterpaper=true,colorlinks,bookmarks=false]{hyperref}

\wacvfinalcopy 

\begin{document}

\title{MURAUER: Mapping Unlabeled Real Data for Label AUstERity}

\author{Georg Poier \hspace{1.5em} Michael Opitz \hspace{1.5em} 
David Schinagl \hspace{1.5em} Horst Bischof\\
Institute for Computer Graphics and Vision\\
Graz University of Technology\\
Austria
}

\maketitle
\thispagestyle{empty}

\blfootnote{\textcopyright \,2019 IEEE}	
\blfootnote{Project webpage providing code and additional material can be found 
at \url{https://poier.github.io/murauer}}

\begin{abstract}
Data labeling for learning 3D hand pose estimation models is a huge effort.
Readily available, accurately labeled synthetic data 
has the potential to reduce the effort.
However, to successfully exploit synthetic data, 
current state-of-the-art methods still require 
a large amount of labeled real data.
In this work, we remove this requirement 
by learning to map from the features of real data 
to the features of synthetic data mainly using 
a large amount of synthetic and \emph{unlabeled} real data.
We exploit unlabeled data using two auxiliary objectives,
which enforce that (i) the mapped representation is pose specific and 
(ii) at the same time, the distributions of real and synthetic data are aligned.
While pose specifity is enforced by a self-supervisory signal requiring that the 
representation is predictive for the appearance from different views,
distributions are aligned by an adversarial term.
In this way, we can significantly improve the results of the 
baseline system, which does not use unlabeled data and 
outperform many recent approaches
already with about 1\% of the labeled real data.
This presents a step towards faster deployment of learning based 
hand pose estimation,
making it accessible for a larger range of applications.

\end{abstract}

\section{Introduction}
To provide labeled data in the needed quantity, accuracy and realism 
for learning pose estimation models currently requires a significant manual effort.
This is especially the case if the goal is to estimate the pose of articulated objects 
like the human hand.
For this task a significant effort has been taken in order to provide 
semi-/automatic labeling procedures and corresponding 
datasets~\cite{Oberweger2016cvpr_anno,Tompson2014tog,Yuan2017cvpr_bighand}.
However, to provide the labeled data for a novel application, viewpoint, 
or sensor still requires significant effort, 
specific hardware and/or great care to not affect the captured data.

\begin{figure}[t]
  \centering
  \includegraphics[width=0.25\linewidth]{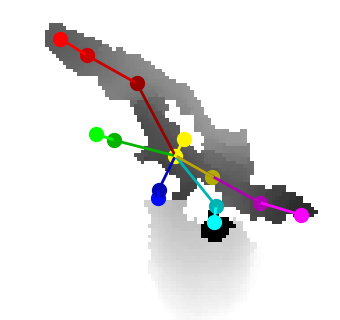}
  \hfill
  \includegraphics[width=0.25\linewidth]{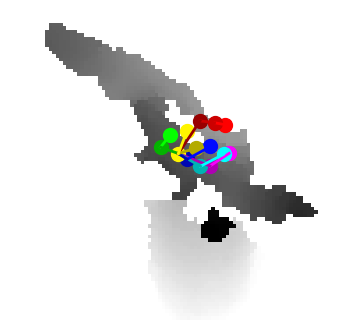}
  \hfill
  \includegraphics[width=0.25\linewidth]{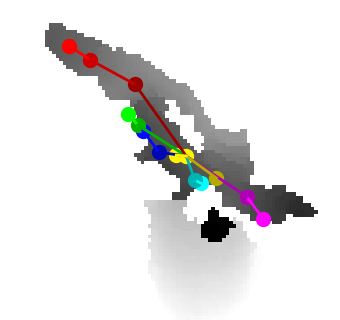}
  \\
  \includegraphics[width=0.25\linewidth]{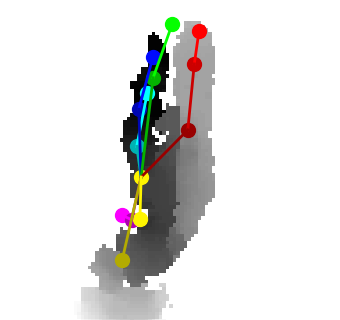}
  \hfill
  \includegraphics[width=0.25\linewidth]{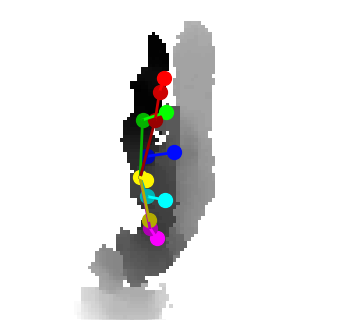}
  \hfill
  \includegraphics[width=0.25\linewidth]{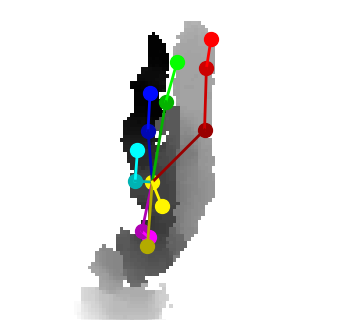}
  \\
  \includegraphics[width=0.25\linewidth]{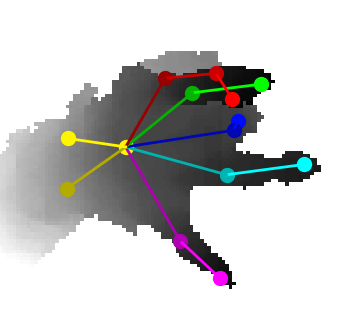}
  \hfill
  \includegraphics[width=0.25\linewidth]{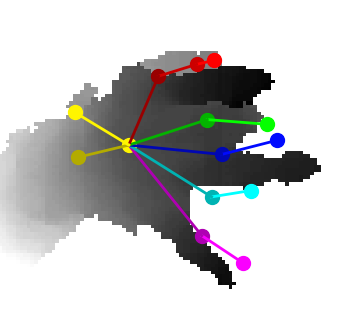}
  \hfill
  \includegraphics[width=0.25\linewidth]{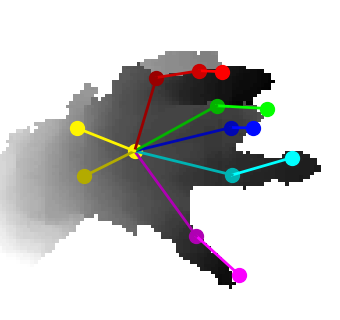}
  \\
  \includegraphics[width=0.25\linewidth]{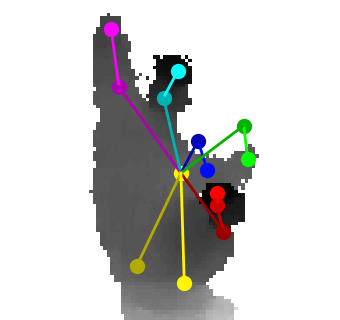}
  \hfill
  \includegraphics[width=0.25\linewidth]{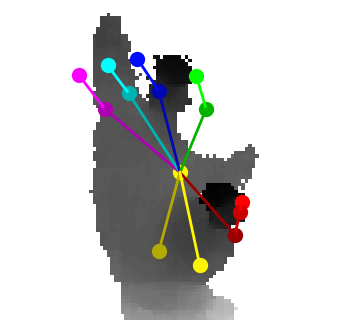}
  \hfill
  \includegraphics[width=0.25\linewidth]{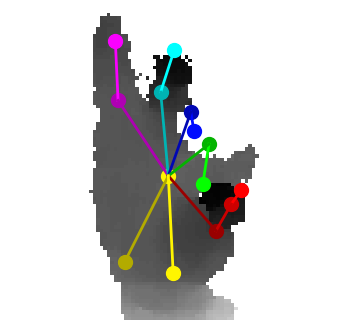}
  \\
  \includegraphics[width=0.25\linewidth]{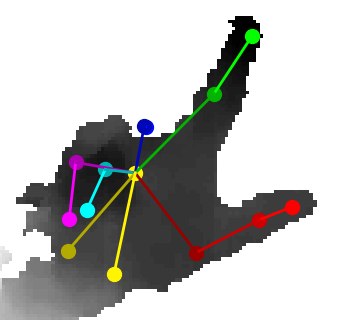}
  \hfill
  \includegraphics[width=0.25\linewidth]{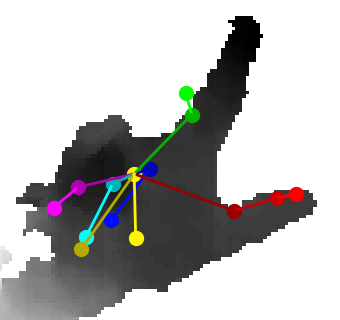}
  \hfill
  \includegraphics[width=0.25\linewidth]{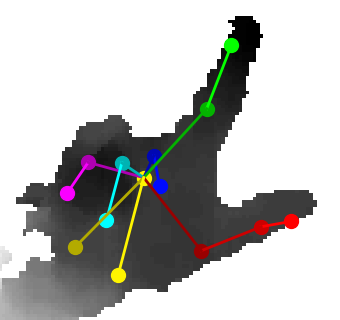}
  \caption[]{\textbf{Comparison of results.} 
  We introduce a method to exploit unlabeled data, which improves results 
  especially for highly distorted images and difficult poses.
  \textbf{Left:} ground truth. \textbf{Middle:} baseline trained with labeled data 
  (synthetic and 100 real). \textbf{Right:} our result from training with the same labeled data 
  and additional unlabeled real data.
  Best viewed in color.
  }
  \label{fig:motivation_figure}
\end{figure}

Recent methods aiming to reduce the effort often employ synthetic data 
or semi-supervised learning~\cite{Neverova2017cviu_weaksupervhape,Wan2017cvpr_crossingnets}, 
which, both, have their specific drawbacks.
Approaches, employing synthetic data have to deal with the domain gap,
which has been recently approached for hand pose estimation by learning a mapping between 
the feature spaces of real and synthetic data~\cite{Rad2018cvpr_featuremap}. 
Unfortunately, learning this mapping 
requires a large amount of labeled real data and corresponding synthetic data.
On the other hand, semi-supervised approaches can better exploit a small amount of 
labeled data, however, the results are often still not 
competitive.

We aim to overcome these issues by exploiting accurately labeled synthetic data
together with unlabeled real data in a specifically devised semi-supervised approach.
We employ a large amount of synthetic data
to learn an accurate pose predictor, 
and, inspired by recent work~\cite{Massa2016cvpr_adapttorendered,Rad2018cvpr_featuremap}, 
learn to map the features of real data to those of synthetic data to overcome 
the domain gap.
However, in contrast to previous work, 
we learn this mapping mainly from unlabeled data.

We train the mapping from the features of real to those of synthetic data using
two auxiliary objectives based on unlabeled data.
One objective enforces the mapped features to be pose 
specific~\cite{Poier2018cvpr_preview,Rhodin2018eccv_viewprediction},
and the other one enforces the feature distributions 
of real and synthetic data to be aligned. 

For the first of the two auxiliary objectives, which is responsible 
for enforcing a pose specific representation, we build upon
our recent work~\cite{Poier2018cvpr_preview}.
In~\cite{Poier2018cvpr_preview} we showed that by learning to predict a different view 
from the latent representation,
the latent representations of similar poses are pushed close together.
That is, the only necessary supervision to learn such a pose specific representation 
can be obtained by simply capturing the scene simultaneously from different view points.
In this work we enforce the joint latent representation of real and synthetic data 
(\ie, after mapping) to be pose specific
by enforcing the representation to be predictive for the appearance in another view.

The second objective is to align the feature distributions of real and synthetic data. 
The underlying idea of learning a mapping from the features of real samples 
to the features of synthetic samples
is that the labeled synthetic data can be better exploited 
if real and synthetic samples with similar poses are  
close together in the latent space.
Simply ensuring that the latent representation is pose specific
does, however, not guarantee that the features of real and synthetic data 
are close together in the latent space:
Similar poses could form clusters for real and synthetic data, \emph{individually}.
To avoid that, we 
employ an adversarial loss, which acts on the latent space and 
penalizes a mismatch of the feature distributions.

By simultaneously ensuring that similar poses are close together 
and feature distributions are aligned, we show that we 
are able to train state-of-the-art pose predictors --
already with small amounts of labeled real data.
More specifically, employing about $1$\% of the labeled real samples from the NYU 
dataset~\cite{Tompson2014tog} our method outperforms many 
recent state-of-the-art approaches, which use all labeled real samples.
Furthermore, besides quantitative experiments,
we perform qualitative analysis showing that the latent representations of real and 
synthetic samples are well aligned when using mainly unlabeled real data. 
Moreover, in our extensive ablation study we 
find that, both, enforcing pose specificity as well as aligning 
the distributions of real and synthetic samples benefits performance 
(see Sec.~\ref{sec:exp:ablation}).

\section{Related work}
Works on hand pose estimation can be categorized into two main approaches.
These two main strands are often denoted \emph{data-driven} and \emph{model-based}.
Data-driven methods~\cite{Keskin2012eccv_multilayeredrf,Madadi2017arxiv_globaltolocal,
Tang2014cvpr_lrf} aim to learn a mapping 
from the input to the pose based on training data and 
thus, crucially, depend on the coverage of the training dataset. 
Model-based approaches~\cite{Gorce2011pami_modelbasedhape,
Oikonomidis2011bmvc,Wu2001iccv_handarticulations}, 
on the other hand,
employ a manually crafted model of the human 
hand and use an analysis-by-synthesis approach, 
where the parameters of the hand-model are optimized
so that the parameterized hand best fits the observed hand.
Traditionally, these approaches relied on an initialization 
of the hand-model parameters from the solution of the previous frame(s) and 
were thus, \eg, subject to tracking errors. 
To address such issues, more recently, these approaches are usually combined with data-driven 
approaches for re-/initialization~\cite{Panteleris2018wacv_hybridmonorgb,Sharp2015chi,
Ye2016eccv_attentionhybridhape}.
However, now the effectiveness of such an approach again
depends on the coverage of the training set
since initialization will not work for out-of-distribution samples.

\paragraph{Training data and annotation}
Given the crucial role of annotated training data for state-of-the-art approaches to 
hand pose estimation, a lot of effort has been devoted to the creation 
of training data sets.
Most often semi-automatic approaches have been employed to label real 
data~\cite{Oberweger2016cvpr_anno,Sun2015cvpr_cascadedhaperegression,Tang2014cvpr_lrf,
Tompson2014tog}.
Still, these are often difficult to set up, require a 
significant amount of manual interaction, 
and the often occurring occlusions make the procedures still difficult or even inaccurate.
In other works, automatic procedures based on attaching 6D magnetic sensors to the 
hand have been developed~\cite{Wetzler2015bmvc_fingertip,Yuan2017cvpr_bighand}.
With these methods, great care has to be taken to avoid that the attached sensors 
affect the data too strongly. 
All these efforts point out 
that the development of methods which reduce the dependence on labeled real data
would foster quicker deployment and make such systems more accessible.

\paragraph{Synthetic data}
One way to lessen the effort for labeling real data is to employ synthetic 
data~\cite{Zimmermann2017iccv_hapefromrgb}.
Synthetic data has the advantage that it has perfectly accurate labels 
and a virtually infinite number of samples can be generated.
However, the data generating distribution usually differs 
between the synthetic training data and the real test data. 
Hence, models trained only on synthetic data suffer from the so-called \emph{domain gap} and 
usually perform significantly worse than models trained on real 
data~\cite{Abdi2018bmvc_lsps,Rad2018cvpr_featuremap}.

\paragraph{Unlabeled data and domain adaptation}
Besides synthetic data, unlabeled real data can be used to lessen the labeling effort.
By relying on the observation 
that the pose is predictive for the appearance of the hand 
seen from another view -- we have recently shown that it is possible 
to learn a pose specific representation using unlabeled data~\cite{Poier2018cvpr_preview}.
In that work~\cite{Poier2018cvpr_preview} we only employ real data and, hence, 
do not have to deal with a domain gap. 
However, for a small number of labeled and a large number of unlabeled data 
this approach alone will not be competitive 
due to the reduced pose supervision.

Other works try to boost performance by combining labeled synthetic and unlabeled real 
data~\cite{Mueller2018cvpr_ganeratedhands,
Shrivastava2017cvpr_synth2realwithadv_applepaper,Abdi2018bmvc_lsps,
Tang2013iccv_semisupervisedhape}. 
To mitigate the domain gap between these two distributions
they typically use a framework based on \acp{GAN}~\cite{Goodfellow2014nips_gan}.
For instance in~\cite{Liu2017arxiv,Mueller2018cvpr_ganeratedhands,
Shrivastava2017cvpr_synth2realwithadv_applepaper} 
a model is learned to transform synthetic images to corresponding real images, 
which can then be used for training using the accurate labels from the initial synthetic data.
Our work is orthogonal. We show how synthetic and unlabeled real data can be used 
to learn a pose specific latent representation, which can directly be used during inference.

Our method is closely related to approaches
that aim to overcome the domain gap by learning a shared latent space 
for different modalities.
For instance in~\cite{Spurr2018cvpr_crossmodallatentspace,Wan2017cvpr_crossingnets} 
a shared latent space is learned for images and poses.
Similarly, Rad~\etal~\cite{Rad2018cvpr_featuremap} incorporate synthetic data 
by learning to map the features of real samples to the features of synthetic samples.
Abdi~\etal~\cite{Abdi2018bmvc_lsps} take the idea of a shared latent space further 
and incorporate poses, synthetic samples as well as labeled and unlabeled real samples.
Similar to~\cite{Wan2017cvpr_crossingnets} they combine a \ac{VAE}~\cite{Kingma2014iclr_vae} 
with a \ac{GAN} 
and exploit unlabeled samples during training the \ac{GAN}, 
which in turn improves the overall system.
In contrast to their work, in our work the adversarial term does not operate on the images
but directly on the much lower dimensional latent space, 
for which it should be easier to train a discriminator-generator pair of lower complexity. 
Moreover, we enforce pose specific constraints on the latent representation of 
both, labeled and unlabeled data,
which yields a significant performance gain (see Sec.~\ref{sec:exp:ablation}).

\section{Semi-supervised feature mapping}
Our method builds on the basic observation that 
for hand pose estimation from depth images
it is easy to obtain labeled synthetic data and unlabeled real data.
First, a large number of synthetic data is used to train a very strong pose predictor.
To make this strong predictor amenable for real data, we learn to map
the real data to synthetic data.
However, we do not want to rely on ground truth supervision to learn this mapping 
between real and synthetic data.
Instead, we learn this mapping by mainly relying on unlabeled data.

\begin{figure*}[t]
\begin{center}
\includegraphics[width=\textwidth]{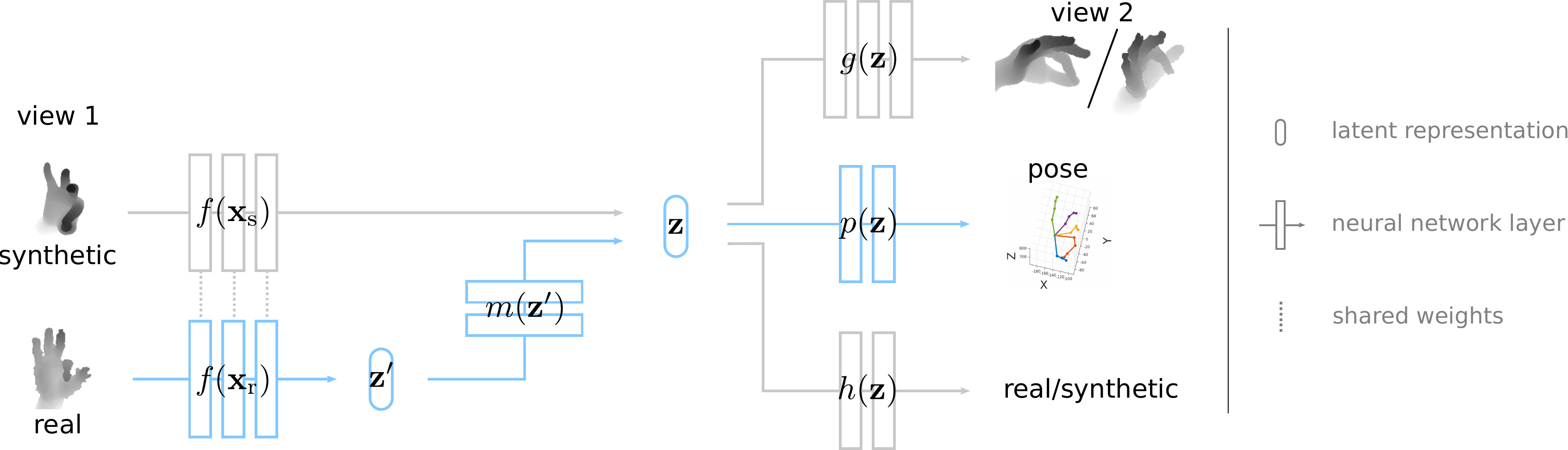}
\end{center}
   \caption{\textbf{Sketch of the architecture.} 
   We train our system jointly with real and synthetic samples.
   From the joint latent representation $\mathbf{z}$, we predict the pose as well as
   two auxiliary outputs from which we also obtain feedback for unlabeled data.
   The auxiliary objectives are (i) to predict a different view and 
   (ii) to discriminate between real and synthetic data.
   The training using unlabeled data ensures aligned latent feature distributions
   and thus, improved exploitation of synthetic data 
   even with a small amount of labeled real samples.
   During test time only the pose is predicted (\textcolor{lightblue_ours}{blue path}).
   The layers per module are just for illustration and 
   do not represent the actual number of layers.
   }
\label{fig:architecture}
\end{figure*}

To properly exploit unlabeled data, we propose two auxiliary loss functions.
The first one uses a self-supervised term,
which enforces the joint latent representation of synthetic and real data to be pose specific 
without the need for pose labels by ensuring that the representation
is predictive for the hands' appearance in another view.
The second loss is an adversarial loss which
ensures that the feature distributions for real and synthetic data are aligned.
That is, we simultaneously ensure that the distributions are matched 
and the representation is pose specific.
Ultimately, the training loss joins the target loss, $\ell_p$, 
with an loss for matching corresponding real and synthetic samples $\ell_{c}$ for labeled data 
and the two auxiliary losses $\ell_{g}$ and $\ell_{m}$, which can also exploit unlabeled data:
\begin{equation}
 \ell = \ell_{p} + \lambda_{c} \ell_{c} + \lambda_{g} \ell_{g} + \lambda_{m} \ell_{m},
 \label{eq:fullloss}
\end{equation}
where $\ell_{g}$ is the loss of the self-supervised term,  
$\ell_{m}$ is the adversarial loss to match the feature distributions, 
and $\lambda_{c}, \lambda_{g}$ and $\lambda_{m}$ are respective weighting terms.
Figure~\ref{fig:architecture} depicts the overall architecture of our method, 
giving rise to the individual loss terms.
We describe all terms in the sequel.

\subsection{Predicting the pose}
For this work we assume the learned model employed for pose prediction 
to be based upon two separate functions.
A function $f$, which transforms the input to some latent space
and a second function $p$, which maps from the latent space to the desired target space.
That is, given an input image $\mathbf{x}$, 
the function $f$ will produce a latent representation:
\begin{equation}
 \mathbf{z} = f(\mathbf{x}).
\end{equation}
The function $p$, on the other hand, maps a given latent representation $\mathbf{z}$ to 
a pose representation,
\begin{equation}
 \mathbf{\hat{y}} = p(\mathbf{z}),
\end{equation}
where the target space can be any pose representation 
(\eg, joint positions). 
Hence -- successively applied -- these two functions map the input image 
to a pose representation:
\begin{equation}
 \mathbf{\hat{y}} = p(f(\mathbf{x})).
\end{equation}

We implemented these two functions as neural networks.
Similar to other works~\cite{Rad2018cvpr_featuremap,Wan2017cvpr_crossingnets}, 
we train the networks to directly output 3D joint positions, 
and use the mean squared error as the loss function to learn the network parameters.
That is, the target loss is simply the squared $L^{2}$ norm:
\begin{equation}
 \ell_{p} = \sum_{k} \left\| \mathbf{y}_{k} - \mathbf{\hat{y}}_{k} \right\|^{2}_{2},
\end{equation}
where $\mathbf{y}_{k}$ is the ground truth and $\mathbf{\hat{y}}_{k}$ is the prediction 
for the $k$-th sample.

\subsection{Mapping real to synthetic data}
To train a neural network for hand pose estimation 
we can generate a virtually infinite amount of synthetic data.
The model trained solely on synthetic data will, 
however, not work similarly well on real images.
To overcome this problem, we take inspiration from recent 
works~\cite{Massa2016cvpr_adapttorendered,Rad2018cvpr_featuremap},
which learn to map the features of real images 
to the feature space of synthetic images.
In this way, a large amount of synthetic images can be exploited 
to train a strong pose prediction model, which then 
-- after mapping the features -- also yields improved performance on real images.

More specifically, a function $m$ is trained to map from the 
features~$\mathbf{z}^{\prime}$ of a real image 
to the feature space of synthetic images:
\begin{equation}
 \mathbf{\hat{z}} = m( \mathbf{z}^{\prime} ),
 \label{equ:mapping}
\end{equation}
where $\mathbf{\hat{z}}$ denotes the latent representation 
of a real image in the feature space of synthetic images.
Hence, employing the whole model to predict the pose of the hand in a real image, 
$\mathbf{x}_{\text{r}}$,
we successively apply functions $f$ to extract features, 
$m$ to map the features and, finally, $p$ to predict the pose:
\begin{equation}
 \mathbf{\hat{y}} = p(m(f( \mathbf{x}_{\text{r}} ))).
\end{equation}

To learn the mapping function $m$,
in~\cite{Massa2016cvpr_adapttorendered,Rad2018cvpr_featuremap}
a one-to-one correspondence between real and synthetic data is required. 
The mapping is then trained to minimize the distance between the mapped 
feature representation $\mathbf{\hat{z}}$ of a real image
and the feature representation $\mathbf{z}$ of the corresponding 
synthetic image.
For the available corresponding real and synthetic samples we follow this approach:
\begin{equation}
 \ell_{c} = \sum_{k \in \mathcal{C}} \left\| \mathbf{z}_{k} 
	    - \mathbf{\hat{z}}_{k} \right\|_{2}^{2},
\end{equation}
where $\mathcal{C}$ denotes the set of available corresponding real and synthetic samples.
However, finding a synthetic image, which accurately corresponds 
to a given real image is indeed equivalent to labeling the real image.
Hence, relying solely on this approach would still require to have a significant amount 
of labeled real images. 
In this work we investigate ways to overcome this requirement 
and reduce the number of necessary corresponding real and synthetic images.
We describe them in the following.

\subsection{Learning to map from unlabeled data}
\label{sec:met:map_unlabeled}
We aim to train the mapping (Eq.~\eqref{equ:mapping})
without requiring a large amount of labeled real samples.
To do this we add two auxiliary loss functions exploiting unlabeled data to train the mapping.
One of them enforces the mapped representation to resemble the pose,
for both, real and synthetic data.
At the same time, a second loss ensures that 
the feature distributions of real and synthetic images 
are not split apart, \ie, pushed away from each other.
Together, these two auxiliary loss functions 
enable us to effectively train the mapping from mainly unlabeled samples.

\subsubsection*{Learning pose specifity from unlabeled data}
We want to map representations of images showing a similar pose
close together.
Since we do not have labels 
we cannot enforce this directly.
Recently, however, we showed that we can learn to generate representations, 
which are very specific to the pose, without having any pose labels~\cite{Poier2018cvpr_preview}.
Here we build upon this idea, which
is based on the observation that the hand pose is predictive for the appearance 
of the hand seen from any known view.
We can exploit this observation using an encoder-decoder model trained to 
predict different views of the hand.
Given the input image from one view,
the encoder of the model infers a latent representation.
The decoder is given the latent representation and trained to predict a second view of the hand.
The idea is that, if the decoder is able to predict another view of the hand
solely from the latent representation, the latent representation must 
contain pose specific information.

That is, we want to learn a decoder $g$, which 
-- given a pose specific feature representation $\mathbf{z}$ -- 
should predict the hands' appearance $\mathbf{x}^{(j)}$ from a different view $j$:
\begin{equation}
 \mathbf{\hat{x}}^{(j)} = g( \mathbf{z}^{(i)} ),
\end{equation}
where $\mathbf{z}^{(i)}$ denotes the feature representation produced 
for an input from view $i$.

To learn such a generator function $g$ we do not need any labels,
we only need to capture the hand simultaneously from a different viewpoint 
or render the hand model from a different virtual view for synthetic samples, respectively.
That is, the objective for the generator -- given only the latent representation -- 
is to predict the appearance of the hand as captured/rendered from the second view.
Hence, to train $g$ we employ a reconstruction loss:
\begin{equation}
 \ell_{g} = \sum_{k} \left\| \mathbf{x}_{k}^{(j)} - \mathbf{\hat{x}}_{k}^{(j)} \right\|_{1},
 \label{equ:previewobjective}
\end{equation}
where $\mathbf{x}_{k}^{(j)}$ is the captured image and 
$\mathbf{\hat{x}}_{k}^{(j)}$ is the model prediction for the $k$-th image from view $j$.

In~\cite{Poier2018cvpr_preview} we showed that only using the view prediction objective, 
the latent representation 
can be enforced to be pose specific without the need for any pose labels.
Nevertheless, for our case we do not have corresponding real and synthetic data.
That is, given a synthetic sample 
the target for the generator $g$ is a synthetic sample 
and, equivalently, for a real sample 
the target for the generator is a real sample. 
In this way the generator $g$ -- besides trying to generate the correct appearance 
corresponding to the pose of the sample -- 
might also try to discriminate between real and synthetic samples in order to accurately
predict the appearance.
This would clearly be counterproductive when aiming to exploit synthetic samples 
and trying to learn a shared latent space of real and synthetic samples.
In the next section we will show how we overcome this issue.

\subsubsection*{Matching feature distributions from unlabeled data}
Enforcing the latent representation $\mathbf{z}$ 
to be specific for the pose does not ensure that real and synthetic samples
with similar poses are mapped to similar latent representations.
Indeed, real and synthetic samples could be pushed into different areas of the feature space 
by the non-linear functions $f$ and $m$, respectively.
Such a separation in the feature space would clearly hamper 
the exploitation of synthetic data for training a pose predictor for real data.

To avoid a scenario where the latent representations of real and synthetic samples 
are pose specific but still separated in the feature space, 
we need a way to ensure that similar poses are mapped to similar latent representations,
independently of whether the samples are real or synthetic.
Without having corresponding real and synthetic samples, 
this is difficult to ensure on the level of individual samples.
However, as long as we can assume that the distribution of poses is similar 
for real and synthetic data,
we can enforce that also the feature distributions match.
By ensuring that the feature distributions match, and 
at the same time ensuring that the features are pose specific,
similar poses should yield similar pose representations for, both, 
real and synthetic samples, which was the initial goal.

Here, we enforce the feature distributions of real and synthetic data
to match by employing an adversarial training loss~\cite{Goodfellow2014nips_gan}.
The adversarial loss operates on the latent representations, 
\ie, we use a discriminator, which is trained to discern real and synthetic 
samples given the latent representation.
The mapping function $m$, on the other hand, should make the latent representation 
of real samples as similar as possible to the latent representation of synthetic 
samples and, hence, indiscernible for the discriminator.

Similar to the formulation in Least Squares GAN~\cite{Mao2017iccv_lsgan}, 
the discriminator function $h$ predicts a real-valued label:
\begin{equation}
 \hat{l} = h( \mathbf{z} ), \quad \hat{l} \in \mathbb{R},
\end{equation}
which should be $l_{\text{r}} = 1$ for real and $l_{\text{s}} = 0$ for synthetic samples,
respectively.
Consequently, the loss for the discriminator penalizes deviations from these target values
for predictions on respective samples:
\begin{equation}
 \ell_{h} = \frac{1}{2} \sum_{k \in \mathcal{R}} \left( \hat{l}_{k} - l_{\text{r}} \right)^{2}
	  + \frac{1}{2} \sum_{k \in \mathcal{S}} \left( \hat{l}_{k} - l_{\text{s}} \right)^{2},
\end{equation}
where $\mathcal{R}$ is the set of real, and 
$\mathcal{S}$ the set of synthetic samples, respectively.

The loss for the mapping function $m$, on the other hand, 
enforces real samples to be indiscernible from synthetic samples for the discriminator:
\begin{equation}
 \ell_{m} = \frac{1}{2} \sum_{k \in \mathcal{R}} \left( \hat{l}_{k} - l_{\text{s}} \right)^{2}.
 \label{equ:distrmatchobjective}
\end{equation}
Our analysis in Sec.~\ref{sec:exp:latentspaceanalysis} shows that in this way the 
latent representations of real and synthetic samples 
can be well aligned from mainly unlabeled samples.

\section{Experiments}
In this section we verify the applicability of our method.

\subsection{Experimental setup}
\label{sec:exp:setup}

In principle, our contribution is agnostic to the network architecture. 
To verify our method we adopt architectures from recent work 
for the individual modules of our system.
We describe the full architecture for $f$, $m$, $g$, $h$ and $p$ 
in the appendix (Sec.~\ref{sec:app:experimentalsetup}) and 
make the implementation of our method publicly 
available\footnote{\url{https://poier.github.io/murauer}}.

As in many recent works we crop a square region around the hand location, 
resize it to a $128\times128$ patch and 
normalize the depth values to the range $[-1,1]$~\cite{Abdi2018bmvc_lsps,
Oberweger2017iccvw_deeppriorpp,Rad2018cvpr_featuremap}.
These patches are used to train the network with a batch size of 64. 
We pre-train $f$ and $p$ with synthetic data for about 170k iterations and 
subsequently train the whole model (Eq.~\eqref{eq:fullloss}) 
jointly with real and synthetic data for 136k iterations. 
We train using \emph{Adam}~\cite{Kingma2015iclr_adam} 
with a learning rate of $3.3 \times 10^{-4}$. 
We use a learning rate warm-up scheme~\cite{Goyal2017arxiv_lrwarmup}, 
which has been shown to work well independent of the batch size, and, 
after 5k iterations, we start to decay the learning rate gradually. 
Again, details can be found in the appendix.

\paragraph{Dataset and metric} 
We employ the NYU hand pose dataset~\cite{Tompson2014tog}, 
since it is the single prominent dataset providing data captured from multiple view points
together with synthetic data, which we can readily use to compare 
to the results of state-of-the-art approaches.
This dataset was captured with three RGBD cameras simultaneously. 
It contains 72,757 frames for training and 8,252 frames for testing.
The validation set used for analyzing the latent space in Sec.~\ref{sec:exp:latentspaceanalysis} 
is a subset of 2,440 samples from the test set 
(as in~\cite{Poier2018cvpr_preview,Wan2017cvpr_crossingnets}).
Following standard convention we evaluate on 14 joints~\cite{Ge2018cvpr_handpointnet,
Moon2018cvpr_v2v,Tompson2014tog} using the commonly used mean joint 
error (ME)~\cite{Moon2018cvpr_v2v,Oikonomidis11iccv,Sun2015cvpr_cascadedhaperegression}. 
The dataset provides a rendered synthetic depth frame corresponding to each of the real images.
While we sample the real images only from the frontal camera, which 
is used for the standard training set, we follow~\cite{Rad2018cvpr_featuremap} 
for our synthetic data set and use images, rendered from the viewpoints of each of the 
three cameras.
That is, we use all 218,271 synthetic samples provided with the dataset.
Note, for the distribution matching loss, 
we sample real and synthetic data only from the 72,757 samples from the frontal view. 
In any case, we randomly transform (details in Sec.~\ref{sec:app:experimentalsetup}) 
every loaded sample individually to increase 
the variability of the data.

In~\cite{Poier2018cvpr_preview} we performed the view prediction experiments 
on a subset of the dataset
since the camera setup was changed during capturing the dataset.
During this work, we found that for a camera pose from before the change of the setup, 
there is a roughly corresponding camera after the change.
Hence, we can use a roughly consistent camera setup for all samples.
In Sec.~\ref{sec:app:previewonfullnyu} we verify that the view prediction 
is not strongly affected by the small camera pose changes across the dataset.

\subsection{Comparison to state-of-the-art approaches}
Here we compare to recent semi-supervised, as well as fully supervised approaches. 

\begin{figure}[t]
  \centering
  \includegraphics[width=\linewidth]{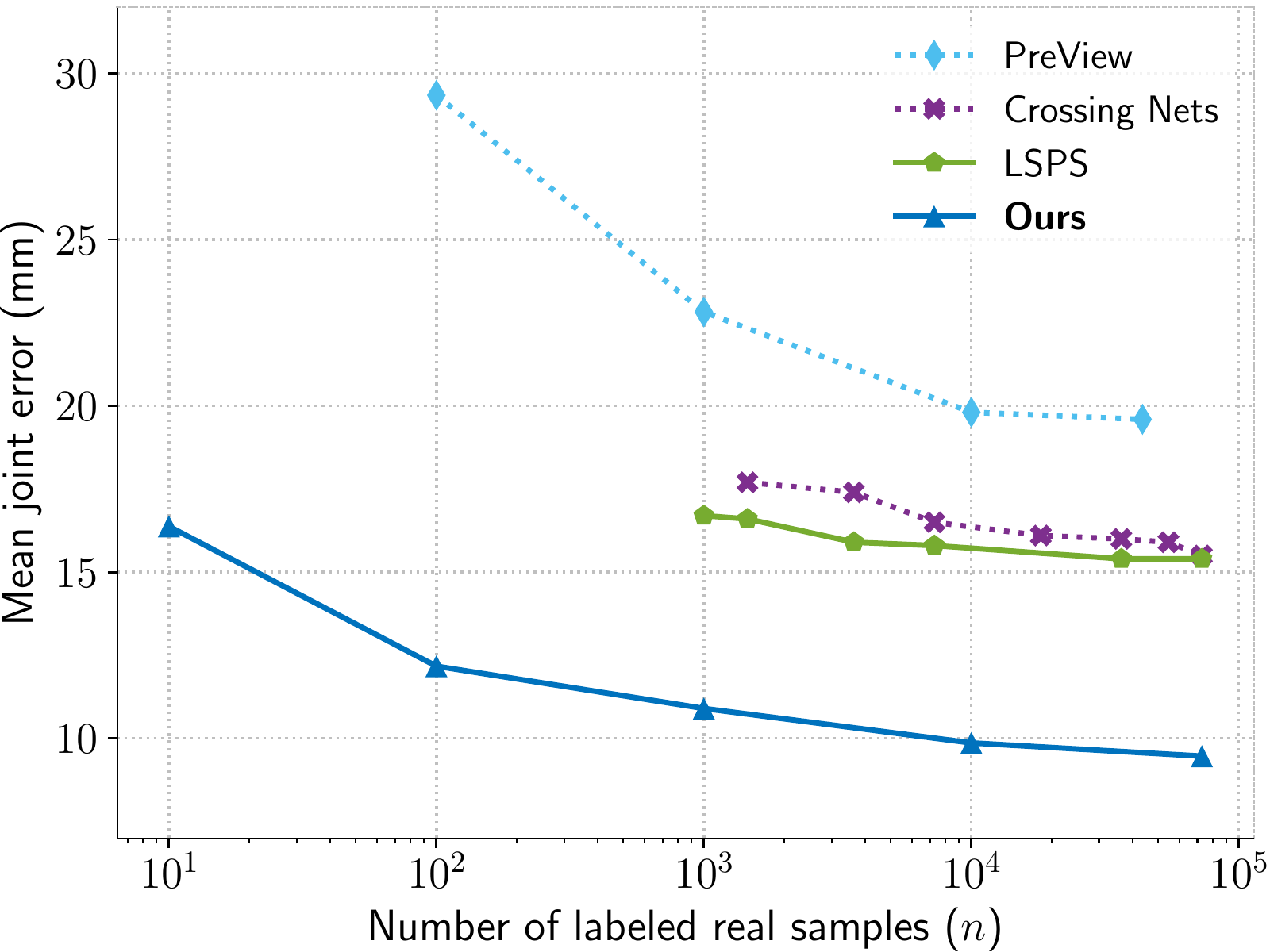}
  \caption[]{\textbf{Comparison to semi-supervised approaches.} 
  Comparison to the recent approaches \emph{PreView}~\cite{Poier2018cvpr_preview}, 
  \emph{Crossing Nets}~\cite{Wan2017cvpr_crossingnets} and \emph{LSPS}~\cite{Abdi2018bmvc_lsps}
  for different numbers of labeled real samples $n$.}
  \label{fig:sota_semisupervised}
\end{figure}

\paragraph{Comparison to semi-supervised methods}
Only a few approaches have recently targeted the semi-supervised setting for 
hand pose estimation: we compare to \emph{Crossing Nets}~\cite{Wan2017cvpr_crossingnets},
\emph{PreView}~\cite{Poier2018cvpr_preview} and \emph{LSPS}~\cite{Abdi2018bmvc_lsps}.
Fig.~\ref{fig:sota_semisupervised} shows the results for different numbers 
of labeled real samples.
Note, that only \emph{LSPS}~\cite{Abdi2018bmvc_lsps} exploits synthetic and real data jointly,
tackling the domain gap.
We compare to the results published by the authors, 
which are provided for different numbers of labeled samples.
Nevertheless, we can see that our method outperforms their results 
independent of the number of labeled real samples.

\paragraph{Comparison on full dataset}
We compare to fully supervised state-of-the-art approaches when employing all labeled data.
We want to stress that our work does not focus on the case where a huge number of 
labeled real samples, roughly covering the space of poses in the test set, is readily available.
We show this comparison, rather, to prove the competitiveness of 
our implementation. 
Tab.~\ref{tab:sota} shows the comparison including some of our baselines.
We can see that the results of our system are within the top state-of-the-art approaches.
Comparing Tab.~\ref{tab:sota} with Fig.~\ref{fig:sota_semisupervised}, 
we see that the results of our method 
are in the range of recent state-of-the-art approaches
even using only a small fraction of the labeled real samples.
Also note that several of the most recent methods
focus on improved input and/or output 
representations~\cite{Chen2018ieeeaccess_shprnet,Ge2018cvpr_handpointnet,
Moon2018cvpr_v2v,Wan2018cvpr_dense3dregr},
which are orthogonal to our work. 

\begin{table}[t]
\begin{center}
\begin{tabular}{l S[table-format=2.1]}
\toprule
Method & {ME (mm)} \\ 
\midrule
DISCO Nets~\cite{Bouchacourt2016nips_disconets} {\color{gray_ours}\footnotesize(NIPS 2016)} & 20.7 \\
Crossing Nets~\cite{Wan2017cvpr_crossingnets} {\color{gray_ours}\footnotesize(CVPR 2017)} & 15.5 \\
LSPS~\cite{Abdi2018bmvc_lsps} {\color{gray_ours}\footnotesize(BMVC 2018)} & 15.4 \\
Weak supervision~\cite{Neverova2017cviu_weaksupervhape} {\color{gray_ours}\footnotesize(CVIU 2017)} & 14.8 \\
Lie-X~\cite{Xu2017ijcv_liex} {\color{gray_ours}\footnotesize(IJCV 2017)} & 14.5 \\
3DCNN~\cite{Ge2017cvpr_3dcnn} {\color{gray_ours}\footnotesize(CVPR 2017)} & 14.1 \\
REN-9x6x6~\cite{Wang2018jvci_ren} {\color{gray_ours}\footnotesize(JVCI 2018)} & 12.7 \\
DeepPrior++~\cite{Oberweger2017iccvw_deeppriorpp} {\color{gray_ours}\footnotesize(ICCVw 2017)} & 12.3 \\
Pose Guided REN~\cite{chen2018pose} {\color{gray_ours}\footnotesize(Neurocomputing 2018)} & 11.8 \\
SHPR-Net~\cite{Chen2018ieeeaccess_shprnet} {\color{gray_ours}\footnotesize(IEEE Access 2018)} & 10.8 \\
Hand PointNet~\cite{Ge2018cvpr_handpointnet} {\color{gray_ours}\footnotesize(CVPR 2018)} & 10.5 \\
Dense 3D regression~\cite{Wan2018cvpr_dense3dregr} {\color{gray_ours}\footnotesize(CVPR 2018)} & 10.2 \\
V2V single model~\cite{Moon2018cvpr_v2v} {\color{gray_ours}\footnotesize(CVPR 2018)} & 9.2 \\
V2V ensemble~\cite{Moon2018cvpr_v2v} {\color{gray_ours}\footnotesize(CVPR 2018)} & 8.4 \\
Feature mapping~\cite{Rad2018cvpr_featuremap} {\color{gray_ours}\footnotesize(CVPR 2018)} & 7.4 \\
\midrule
Synthetic only & 21.3 \\
Real only & 14.7 \\
Real and Synthetic & 13.1 \\
\textbf{Ours} & 9.5 \\
\bottomrule
\end{tabular}
\end{center}
\caption{
  \textbf{Comparison to state-of-the-art.}
  Mean joint error (ME) for training with all labeled real samples from the NYU dataset 
  for recent state-of-the-art approaches, baselines and our method.
  }
\label{tab:sota}
\end{table}

The comparisons in this section are based upon the numbers published by the authors.
That is, these comparisons disregard differences in the used 
data subsamples, models, architectures, and other specificities.
For a better evaluation of the contribution of our work
we investigate the different ingredients of our approach 
based on the same experimental setup in the next section.

\subsection{Ablation study}
\label{sec:exp:ablation}

\begin{figure}[t]
  \centering
  \includegraphics[width=\linewidth]{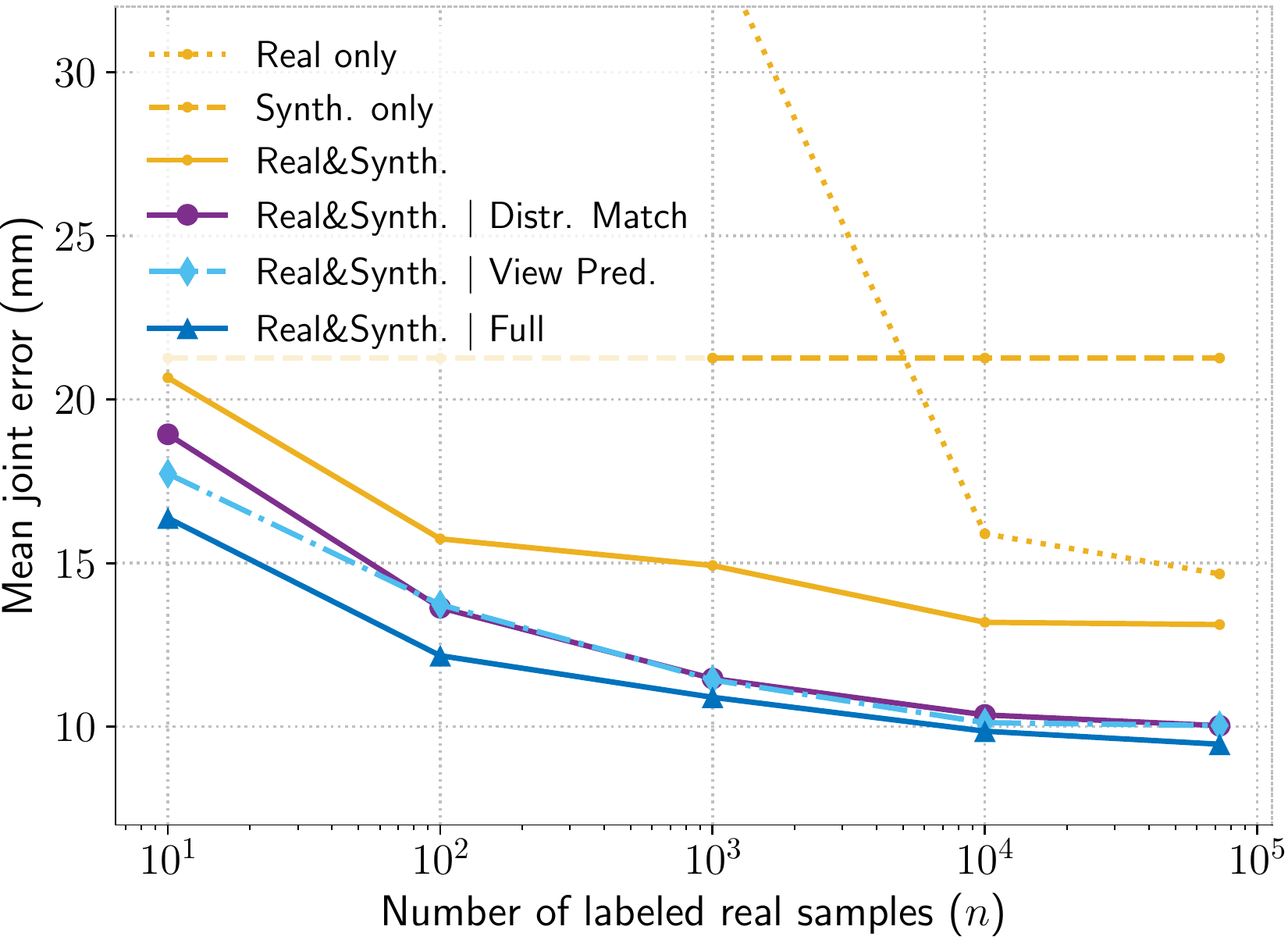}
  \caption[]{\textbf{Ablation experiments.} 
  How different aspects of our work influence the performance
  over different numbers of labeled real samples $n$.
  \emph{Real} and \emph{Synth.} specifies whether real or synthetic data was used and 
  further descriptions identify different variants of our method. See text for details.
  }
  \label{fig:ablationstudy}
\end{figure}

In the ablation study we aim to compare our method to 
baselines based on the same experimental setup and
investigate how effective our contributions are.
To this end, we use the same architecture and train it with different data:
with labeled real only, with synthetic only, with labeled real and synthetic, or 
with labeled real, synthetic and additional unlabeled real data.
Fig.~\ref{fig:ablationstudy} shows the results for different numbers of labeled real samples.
We compare to different variants of our method denoted \emph{Real+Synth. $|$ *}, 
where the asterisk (*) acts as a placeholder for how we train the mapping and thus 
exploit unlabeled data.
That is, we compare the full implementation of our method (\emph{Real\&Synth. $|$ Full}, 
Eq.~\eqref{eq:fullloss}) and two ablated variants: 
One variant where the exploitation of unlabeled data is only based on the adversarial loss term 
(\emph{Real+Synth. $|$ Distr. Match}), and
another variant where only the view prediction objective is used for unlabeled data
(\emph{Real+Synth. $|$ View Pred.}).

We see that each of the individual loss terms yields a significant performance gain 
compared to the baseline system, which uses real and synthetic data but cannot 
exploit unlabeled data.
The additional gain is more enhanced for a small number of labeled real samples $n$ 
and only small for large $n$, but consistent over all $n$.

\subsection{Latent space analysis}
\label{sec:exp:latentspaceanalysis}

Finally, we investigate how the introduced method affects the shared latent space 
of real and synthetic data. 

\begin{figure*}[t]
  \centering
  \includegraphics[width=\textwidth]{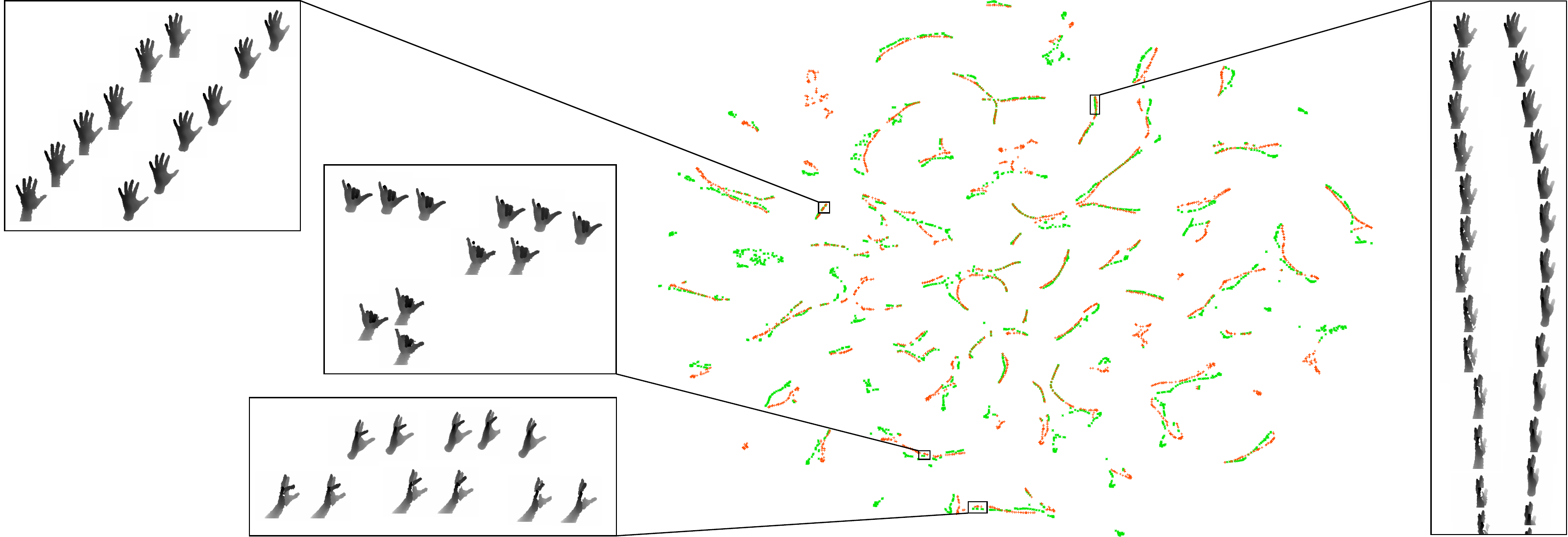}
  \caption[]{\textbf{Visualization of latent representations.} 
  t-SNE visualization of the learned latent representation of 
  \textcolor{greenforest_ours}{real (green; \ding{54})} and 
  \textcolor{orangezinnia_ours}{synthetic (orange; \ding{58})} samples 
  from the validation set.
  Simultaneously, real and synthetic samples as well as similar poses 
  are aligned in the latent representation,
  while only 100 corresponding real and synthetic images are employed during training.
  Note, if necessary, we moved the visualized depth images slightly apart, 
  so, that they do not overlap.
  Best viewed in color with zoom.
  }
  \label{fig:tsne_visualization}
\end{figure*}

\paragraph{Visualization}
We compute the latent representations of corresponding real and synthetic samples 
from the validation set and 
visualize the representations using 
t-SNE~\cite{VanDerMaaten2008jmlr_tsne}.
From the t-SNE visualization in Fig.~\ref{fig:tsne_visualization} 
we can see that the real and synthetic data is well aligned and 
that the aligned data points correspond to similar poses.
This is illustrated by the depth images for exemplary parts of the representation.
Nevertheless, such visualizations have to be interpreted with caution 
(see, \eg, \cite{Wattenberg2016distill_tsne}). 
Hence, we show more visualizations only in the appendix (Sec.~\ref{sec:app:qualitativeanalysis}) 
and try to get more insights from analyzing the distances directly.

\begin{figure}[t]
  \centering
  \includegraphics[width=\linewidth]{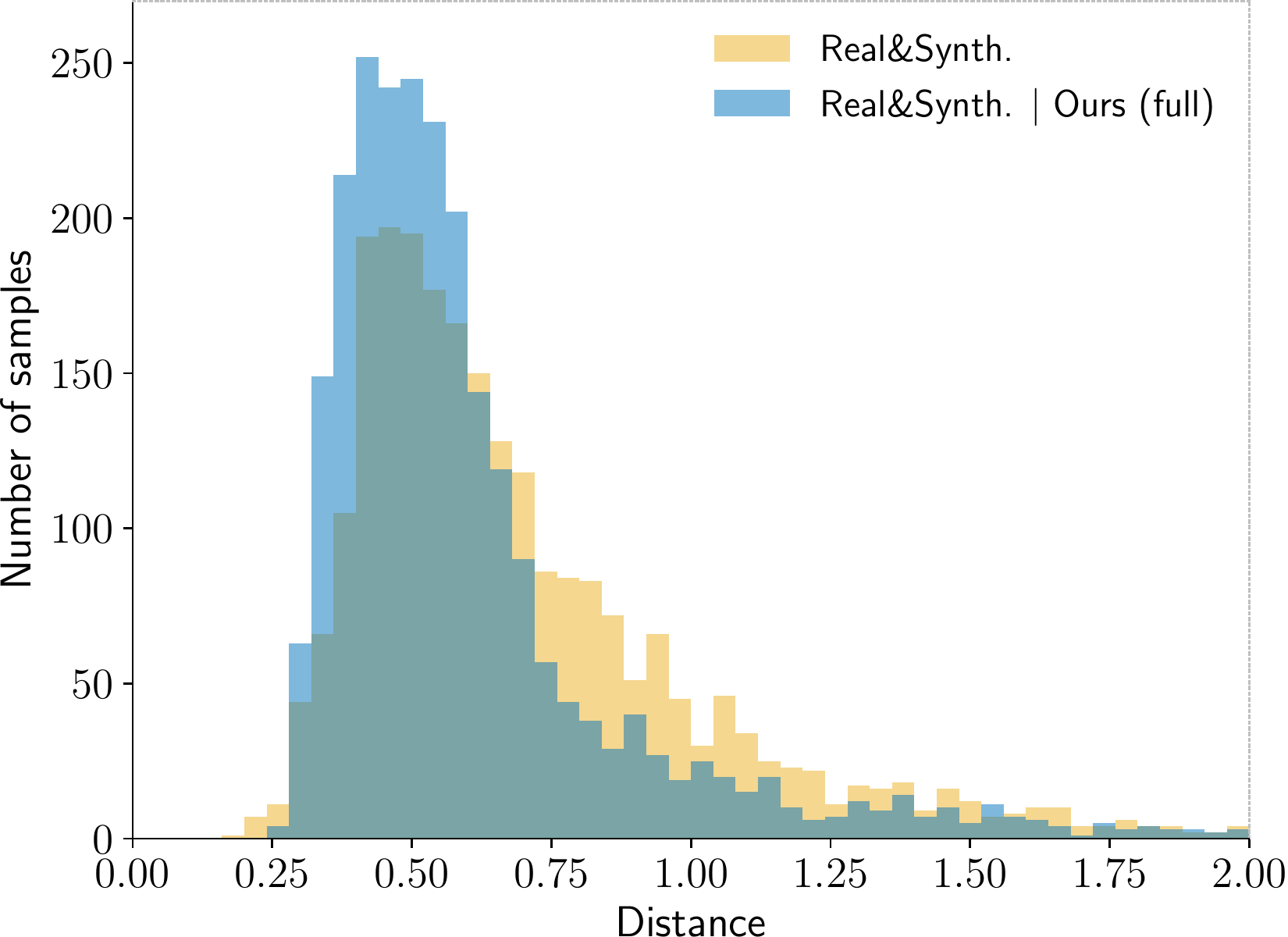}
  \caption[]{\textbf{Distributions of latent space distances between 
  corresponding real and synthetic samples.} 
  Comparison of the distance distributions for 
  \textcolor{lightblue_ours}{our method} and a \textcolor{yellow_ours}{baseline}. 
  The higher peak for lower distances shows that 
  our method moves  corresponding real and synthetic data closer together.
  See text for details.}
  \label{fig:emb_distance_distributions}
\end{figure}

\paragraph{Distance distributions}
To better investigate how our contributions affect the latent space distributions, 
we again make use of the fact that we have corresponding real and synthetic validation samples.
We compute the distances of the latent representations of 
corresponding real and synthetic samples and 
compare the distribution of these distances for different experiments.
In Fig.~\ref{fig:emb_distance_distributions} we compare the distance distribution for:
(i) a baseline experiment which was trained jointly with synthetic and 100 labeled real samples, 
and (ii) our approach, which was trained with the same labeled data but
additionally employs unlabeled data.
We can see, that despite the weak supervision from the unlabeled data, \ie 
correspondence is not known and only the additional loss terms described in 
Sec.~\ref{sec:met:map_unlabeled}
can be used to match the data, the distance between the corresponding 
validation samples are clearly smaller.

\begin{figure}[t]
  \centering
  \includegraphics[width=0.32\linewidth]{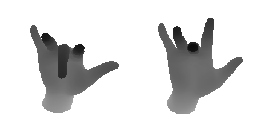}
  \includegraphics[width=0.16\linewidth]{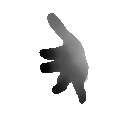}
  \hfill
  \includegraphics[width=0.32\linewidth]{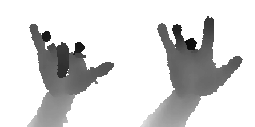}
  \includegraphics[width=0.16\linewidth]{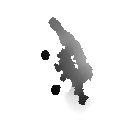}
  \\
  \includegraphics[width=0.32\linewidth]{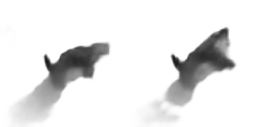}
  \includegraphics[width=0.16\linewidth]{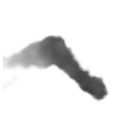}
  \hfill
  \includegraphics[width=0.32\linewidth]{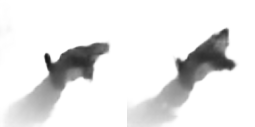}
  \includegraphics[width=0.16\linewidth]{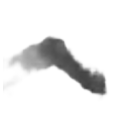}
  \caption[]{\textbf{Example view predictions for real and synthetic input.} 
  \textbf{Top:} Three corresponding synthetic (left) and real (right) 
  validation images.
  \textbf{Bottom:} Predicted views for synthetic (left) and real (right) input.
  }
  \label{fig:viewpredictionsamples}
\end{figure}

\paragraph{Example view predictions}
Finally, we compare examples for view predictions of our method 
given input from either real or synthetic data.
This is interesting, since a possible drawback of the view prediction objective 
is that the generator $g$ 
might try to discriminate between real and synthetic data in order to predict 
the appearance accurately, as has been discussed in Sec.~\ref{sec:met:map_unlabeled}.
However, by looking at the predicted views (\cf, Fig.~\ref{fig:viewpredictionsamples}) 
we see that this is not the case.
We rather find that the predictions are nearly equivalent for real and synthetic samples 
with the same pose, again indicating that similar poses 
are close together in the latent space -- independent of the domain -- 
which was the intention of the contributions introduced in this work.

\section{Conclusions}
In this work we focused on the exploitation of synthetic data 
for the task of 3D hand pose estimation from depth images.
Most importantly, we showed that the existing domain gap between real and synthetic data, 
which hampers the exploitation, can be reduced using mainly unlabeled real data.
To this end, we introduced two auxiliary objectives, which ensured that 
input images exhibiting similar poses are close together in a shared latent space -- 
independent of the domain they are from.
We showed that our method outperforms many recent state-of-the-art approaches
using a surprisingly small fraction of the labeled real samples they use.
We believe that the largely reduced labeling effort renders such systems 
more accessible for a range of applications 
for which this effort has been the limiting factor.

{
\paragraph{Acknowledgements}
We thank the anonymous reviewers for their effort and valuable feedback, and 
Markus Oberweger for his feedback regarding their implementation 
of~\cite{Rad2018cvpr_featuremap}.
}

{\small
\bibliographystyle{ieee}
\bibliography{../_abbrv_short,../references}
}

\clearpage
\pagenumbering{roman}
\setcounter{page}{1}
\appendix

\section{Appendix}
In this appendix we provide more details about our experimental setup and 
experimental results supporting the claims of the paper.
Additionally, we analyze the main error cases.

\subsection{Details for experimental setup}
\label{sec:app:experimentalsetup}

In the following we describe some more details about our experimental setup.
Note that we also make our implementation publicly available\footnote{
\url{https://poier.github.io/murauer}}.

\paragraph{Architecture}
For the network architectures of the individual modules of our work 
(\cf Fig.~\ref{fig:architecture}), 
we relied on architectures which have proven successful in related work:
The feature extractor $f$ is similar to the model used in~\cite{Oberweger2017iccvw_deeppriorpp},
\ie, an initial convolutional layer with 32 filters of size $5\times5$ is followed by 
a $2\times2$ max-pooling, four ``residual modules'' with 64, 128, 256, and 256 filters, 
respectively, 
each with five residual blocks~\cite{He2016cvpr_resnet}, and 
a final fully connected layer with 1024 output units.
The pose estimator $p$ consists of two fully connected layers, with 1024 and $3J$ outputs, 
respectively, where $J$ is the number of predicted joint positions in our case.
The mapping layer $m$ is adopted from~\cite{Rad2018cvpr_featuremap}, 
\ie, it consists of two residual blocks, each with 1024 units.
The discriminator $h$ has the same architecture as the mapping $m$ with an additional 
linear layer to predict a single output.
The generator $g$ uses the architecture of the decoder described in~\cite{Poier2018cvpr_preview},
which is based on the generator of DCGAN. 
It consists of four layers of transposed convolutions, each followed by 
Batch Normalization~\cite{Ioffe2015icml_batchnorm} and a 
leaky ReLU activation~\cite{Maas2013icml_rectifiernonlinearities}.
We add a bilinear upsampling layer prior to the final hyperbolic tangent (tanh) activation
in order to upsample from $64\times64$ to $128\times128$ in our case.

\paragraph{Optimization}
For optimization of the model parameters we use Adam~\cite{Kingma2015iclr_adam} 
with standard parameters, \ie, $\beta_{1} = 0.9$ and $\beta_{2} = 0.999$.
We also found it helpful to follow a \emph{warm-up} scheme 
for the learning rate and 
decay the learning rate gradually later~\cite{Goyal2017arxiv_lrwarmup}.
More specifically, we start with about one tenth of the learning rate and 
approximately triple it after the first epoch.
Subsequently, the learning rate is subject to exponential decay.
That is, the learning rate $\alpha_{e}$ for epoch $e$ is computed by:
\begin{equation}
 \alpha_{e} = \eta_{e} \, \alpha_{0},
\end{equation}
with the scaling factor 
\begin{equation}
 \eta_{e} = 
  \begin{cases}
   0.33^{2-\lfloor \frac{e}{2} \rfloor} 	& \text{if } e < 4 \\
   \exp(-\gamma \, e) 	& \text{otherwise},
  \end{cases}
\end{equation}
where $\gamma$ determines the speed of the decay and is set to $0.04$ in our case.
In our experiments $\alpha_{0} = 3.3 \times 10^{-4}$ yielded the best results.
Here, the notion of epoch is always based on the number of real data samples in the dataset
(72,757 for the NYU dataset) and independent of the actually used dataset (\eg, sub-sampled 
real data, synthetic data, \etc).
That is, the number of iterations per epoch is the same for all experiments 
(1,137 with a batch size of 64).

\paragraph{Loss weights $\lambda$ and mini-batch sampling}
We experimentally found the loss weights used in our work and 
set $\lambda_{c} = 0.2$, $\lambda_{g} = 10^{-4}$ and $\lambda_{m} = 10^{-5}$.
For each mini-batch we independently sample a set of corresponding real and synthetic samples,
a set of real samples, a set of synthetic samples and a set of unlabeled samples 
such that there is an equal number of samples from each of the four sets 
(\ie, 16 samples per set in our case).

\paragraph{Data augmentation} 
We used online data augmentation. That is, each time we sample a specific image 
we also sample new transformation parameters.
In this work we randomly rotate the loaded image, 
randomly sample the location of the crop and 
add white noise to the depth values. 
The rotation angle is uniformly sampled from $[\ang{-60}, \ang{60}]$ and 
the location offset as well as white noise is sampled from a normal distribution 
with $\sigma = \SI{5}{\milli\meter}$.

\subsection{Full NYU dataset for view prediction}
\label{sec:app:previewonfullnyu}
In our recent work~\cite{Poier2018cvpr_preview}, in which we showed that 
the view prediction objective is a good proxy for pose specifity
we decided to leave out about 40\% of the NYU dataset 
for learning to predict different views because the camera setup has been changed 
when capturing this part.
As briefly mentioned in Sec.~\ref{sec:exp:setup}, 
we found that the camera views in the left-out part, 
which are closest to the view points we used for the NYU-CS in~\cite{Poier2018cvpr_preview} 
are roughly the same for at least one camera throughout the whole dataset.
That is, ignoring slight changes of the camera poses, we could employ 
the whole NYU dataset, despite assuming a fixed setup as in~\cite{Poier2018cvpr_preview}.

\paragraph{View prediction}
Here, we show that a model trained for view prediction
can exploit the full set, despite the slight changes of the setup. 
To this end, we train on the reduced NYU-CS set~\cite{Poier2018cvpr_preview} 
or the full NYU dataset, respectively, and compare the results on the standard test set.
In our experiments the \ac{MAE} is reduced by 13\% (see Tab.~\ref{tab:resultsonfullnyu})
when exploiting the full NYU dataset instead of the reduced NYU-CS.
This shows that even for the base task of view prediction 
the additional data can be exploited despite the slightly changed camera poses.

\begin{figure}[t]
\begin{center}
  \includegraphics[width=\linewidth]{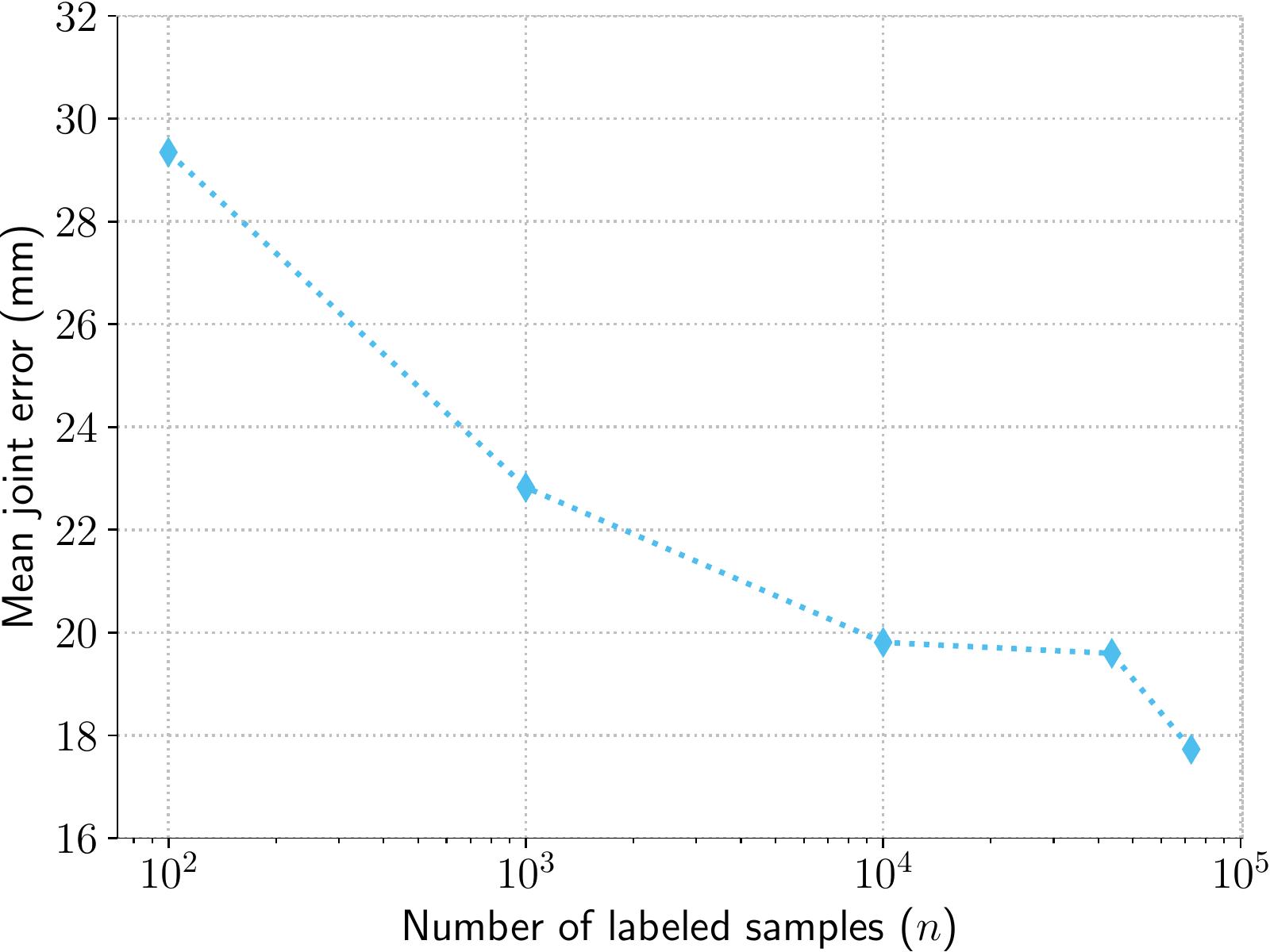}
\end{center}
  \caption{\textbf{Semi-supervised learning with additional data.}
  Results of the semi-supervised method introduced in~\cite{Poier2018cvpr_preview}
  when evaluated on the NYU-CS subset~\cite{Poier2018cvpr_preview} with $\sim$44k samples 
  (\ie, leftmost four experiments in plot) and 
  when exploiting the full NYU set with $\sim$73k samples. See text for details.}
\label{fig:semipreview_withalldata}
\end{figure}

\paragraph{Semi-supervised hand pose estimation}
Furthermore, we evaluate the model we used in~\cite{Poier2018cvpr_preview} 
using the full dataset directly on hand pose estimation. 
We investigate how the error evolves with a gradually increased number of
labeled samples.
We gradually increase the number of labeled subsets of the NYU-CS dataset 
up to the $\sim$44k samples from NYU-CS
and compare to the result when using all ($\sim$73k) samples in 
Fig.~\ref{fig:semipreview_withalldata}.
We can see that the error starts to level up when using subsets from NYU-CS, but experiences 
a sudden drop when exploiting the additional data.
The fact that the results do not just improve gradually
-- as it would probably be the case if we would 
provide ``more of the same'' data, \ie, a denser sampling of the existing data --
again indicates that the model can indeed exploit the additional data included in the full set.

\begin{table}[t]
\begin{center}
\begin{tabular}{l c}
\toprule
Dataset ($n$) 	& \acf{MAE} \\ 
\midrule
NYU-CS (43,640) & 0.123 \\
NYU (72,756) 	& 0.107 \\
\bottomrule
\end{tabular}
\end{center}
\caption{
  \textbf{View prediction with additional data.}
  The results of view prediction 
  trained on the NYU-CS subset as used in~\cite{Poier2018cvpr_preview}
  and on the full NYU set by using the camera views with 
  roughly the same viewpoints for each frame. See text for details.
  }
\label{tab:resultsonfullnyu}
\end{table}

\subsection{Qualitative analysis}
\label{sec:app:qualitativeanalysis}

\begin{figure*}[t]
  \centering
  \begin{subfigure}[b]{0.48\textwidth}
    \includegraphics[width=\textwidth]{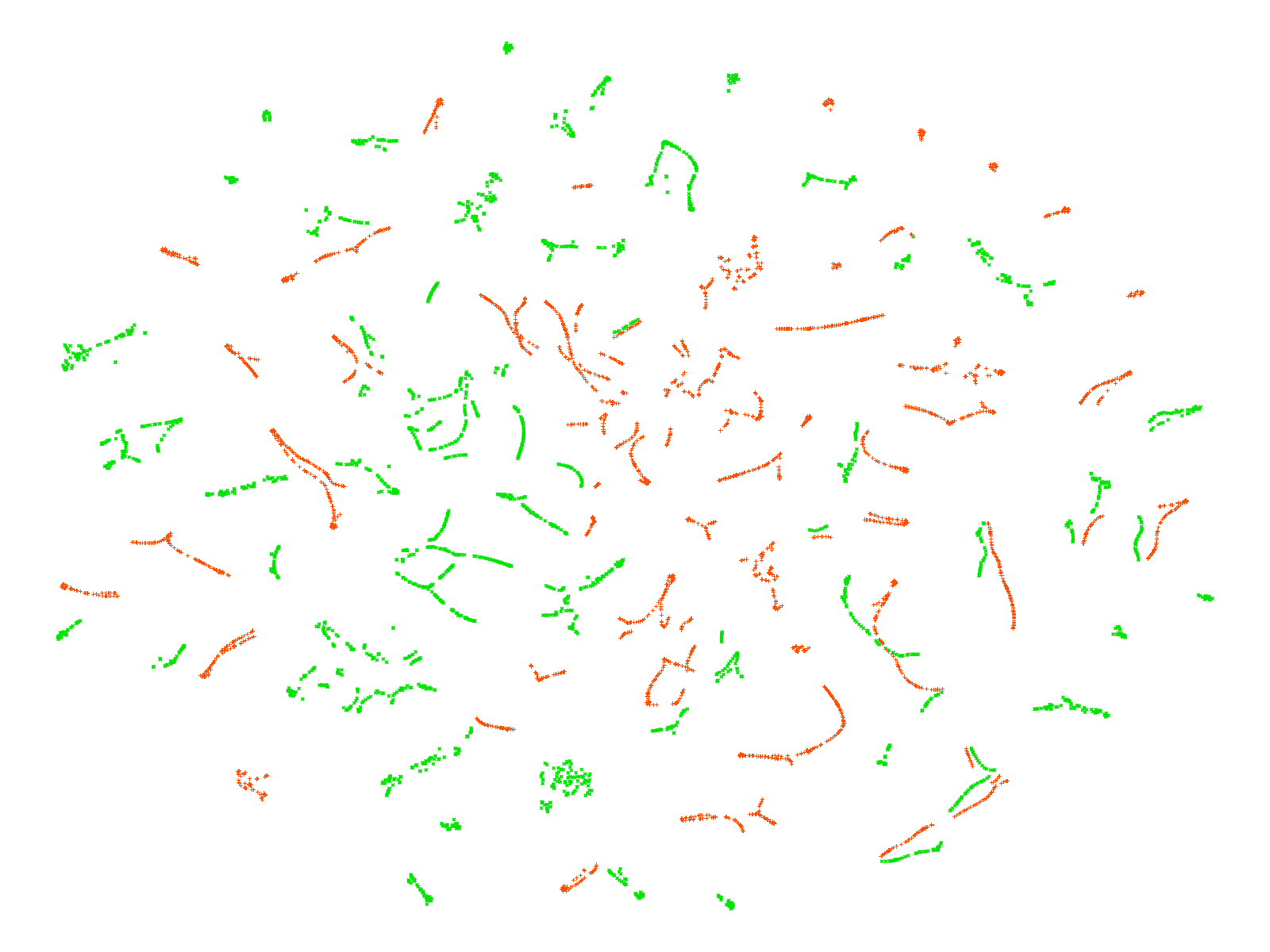}
    \caption{}
    \label{fig:tsne_viz_synthonly}
  \end{subfigure}
  \hfill
  \begin{subfigure}[b]{0.48\textwidth}
    \includegraphics[width=\textwidth]{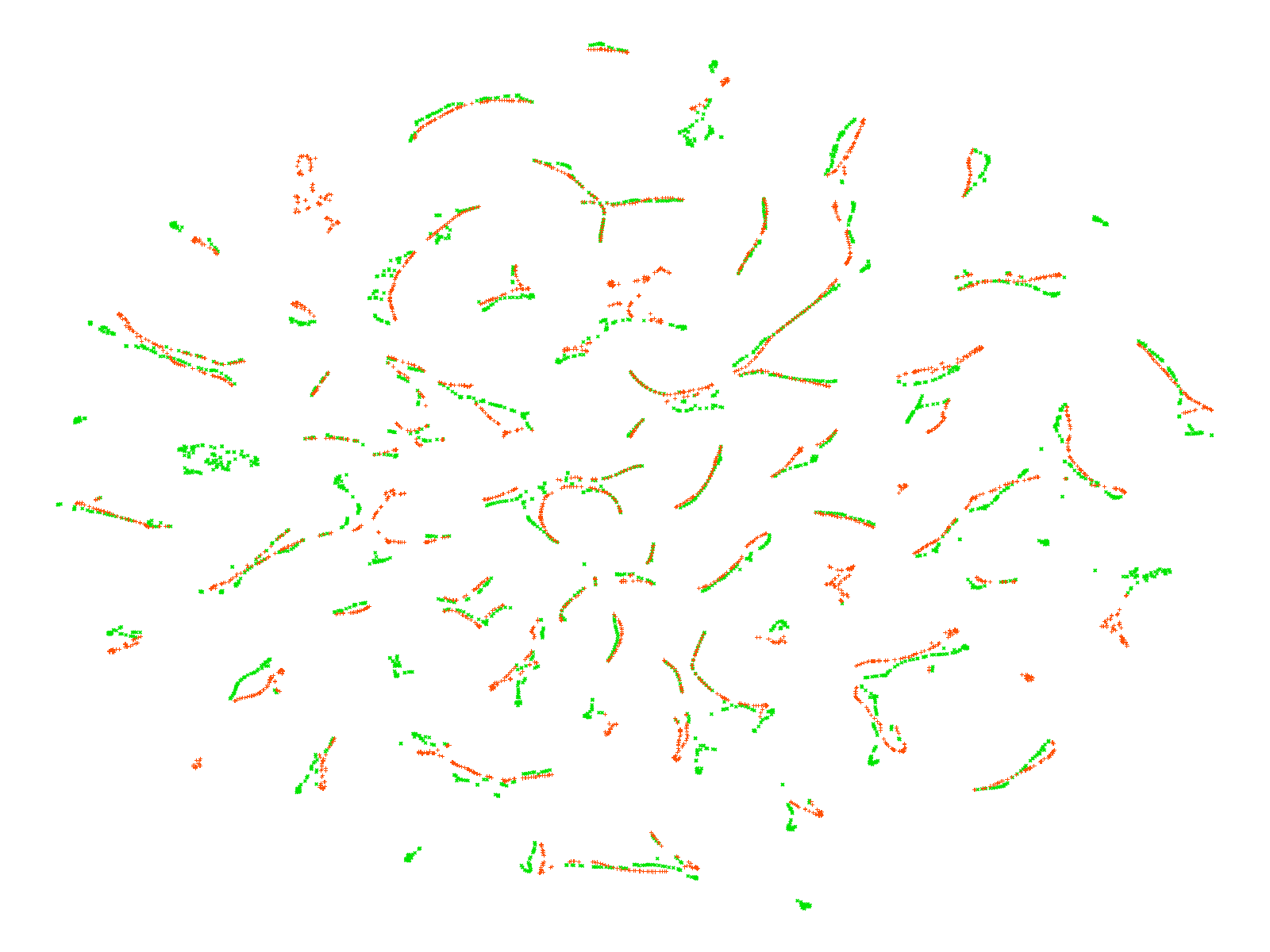}
    \caption{}
    \label{fig:tsne_viz_ours}
  \end{subfigure}
  \caption[]{\textbf{Visualization of latent representations.} 
  The latent representations of corresponding 
  \textcolor{greenforest_ours}{real (green; \ding{54})} and 
  \textcolor{orangezinnia_ours}{synthetic (orange; \ding{58})} samples 
  visualized using t-SNE~\cite{VanDerMaaten2008jmlr_tsne}. 
  Visualization for a model trained on synthetic data (a) and our model 
  trained on synthetic, unlabeled and 0.1\% of labeled real samples (b).
  Best viewed in color.
  }
  \label{fig:tsne_viz_comparision}
\end{figure*}

\paragraph{Domain gap in latent space}
Above, in Fig.~\ref{fig:tsne_visualization} we showed a t-SNE visualization 
of the latent representation learned with our method.
In a similar manner, here, 
we want to qualitatively illustrate the importance of tackling the domain gap between 
synthetic and real data when exploiting synthetic data for hand pose estimation.
To this end, we compare the t-SNE visualization of the latent space learned solely
with synthetic data to the the visualization of the representation obtained with our method 
in Fig.~\ref{fig:tsne_viz_comparision}.
Despite that the poses of real and synthetic images
are corresponding, we can see that the samples from the two
domains take up rather different areas in the visualization
for the model trained only on synthetic data (see Fig.~\ref{fig:tsne_viz_synthonly}).
Whereas, the visualization for our method, which also uses unlabeled data and 
only 100 labeled real samples, shows that the
real and synthetic data is well aligned (see Fig.~\ref{fig:tsne_viz_ours}).

\begin{figure*}[t]
  \centering
  \includegraphics[width=0.19\linewidth]{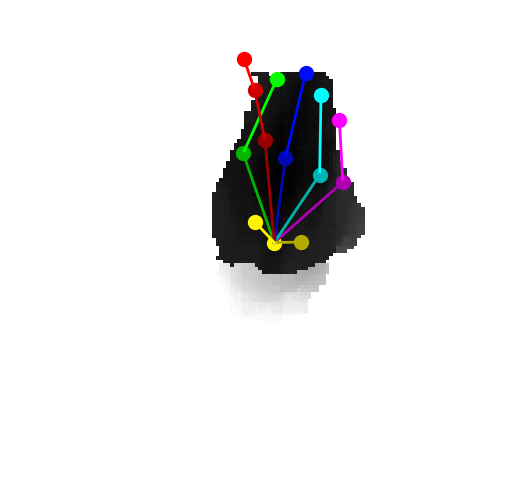}
  \hfill
  \includegraphics[width=0.19\linewidth]{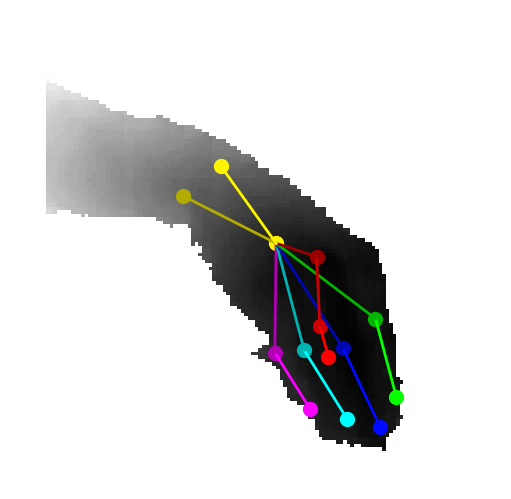}
  \hfill
  \includegraphics[width=0.19\linewidth]{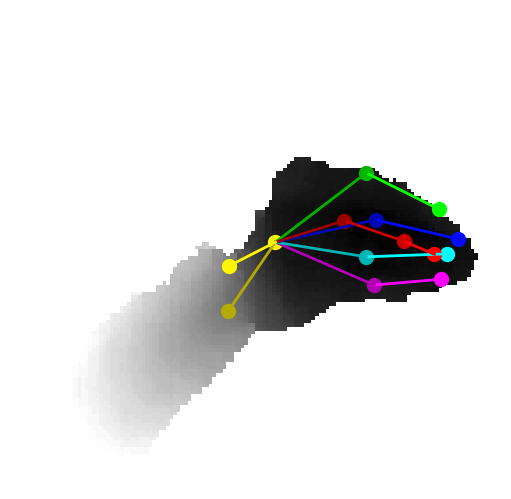}
  \hfill
  \includegraphics[width=0.19\linewidth]{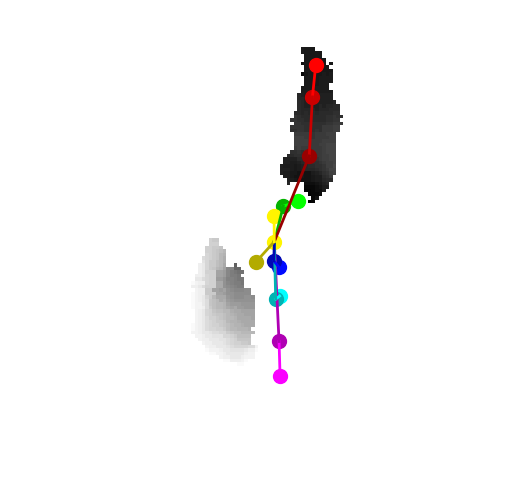}
  \hfill
  \includegraphics[width=0.19\linewidth]{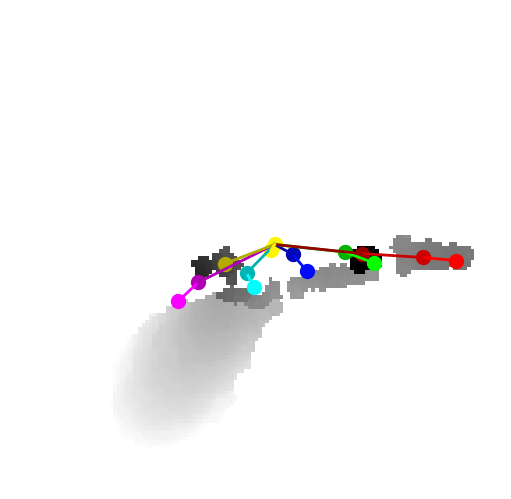}
  \\
  \includegraphics[width=0.19\linewidth]{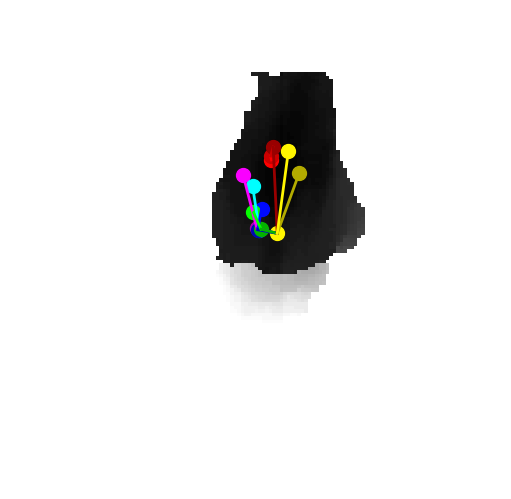}
  \hfill
  \includegraphics[width=0.19\linewidth]{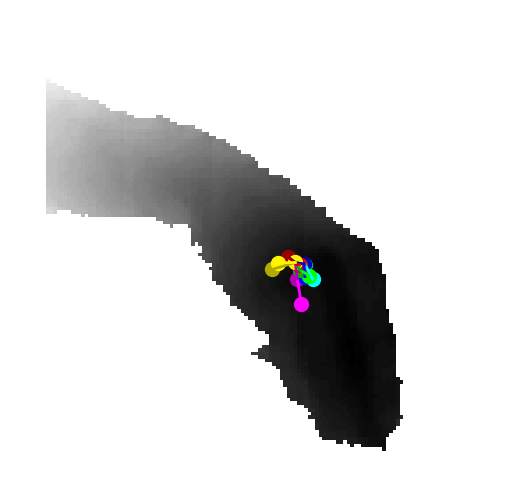}
  \hfill
  \includegraphics[width=0.19\linewidth]{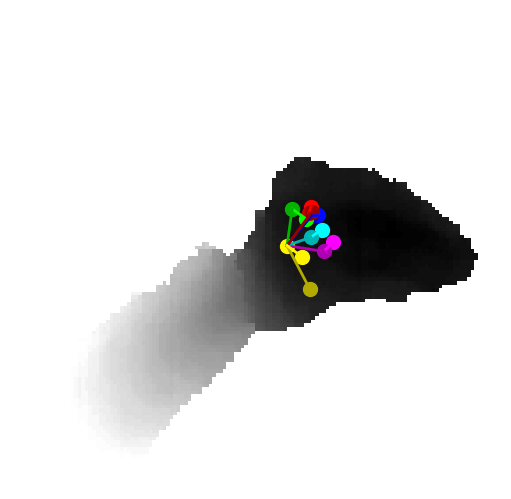}
  \hfill
  \includegraphics[width=0.19\linewidth]{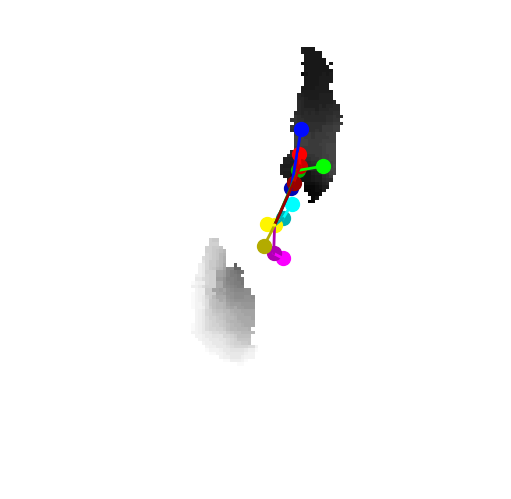}
  \hfill
  \includegraphics[width=0.19\linewidth]{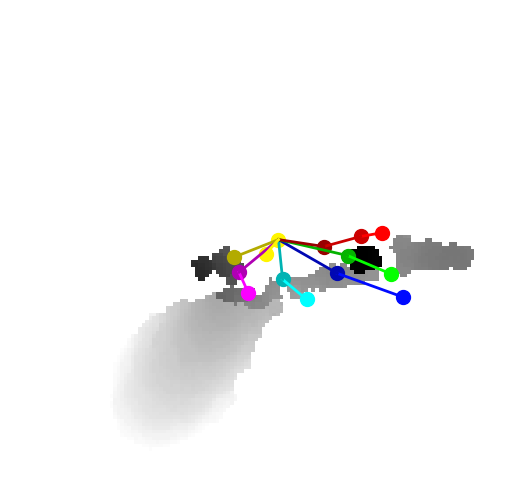}
  \caption[]{\textbf{Frames with largest mean error.} 
  Test samples overlaid with ground truth (top row) and 
  the predictions of our model (bottom row).
  Note, 90 of the 100 frames with the largest mean error are 
  variations of the leftmost three frames.
  }
  \label{fig:large_meanerror_frames}
\end{figure*}

\begin{figure*}[t]
  \centering
  \includegraphics[width=0.19\linewidth]{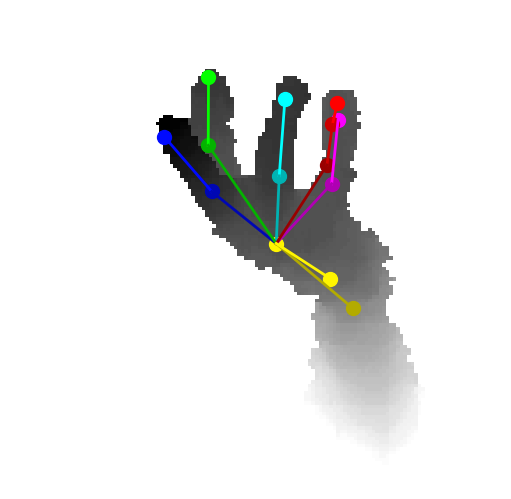}
  \hfill
  \includegraphics[width=0.19\linewidth]{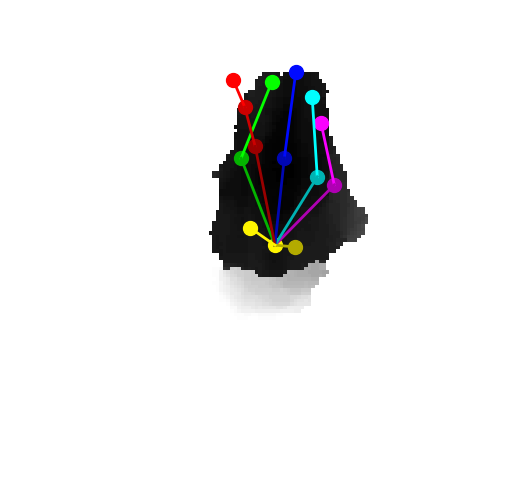}
  \hfill
  \includegraphics[width=0.19\linewidth]{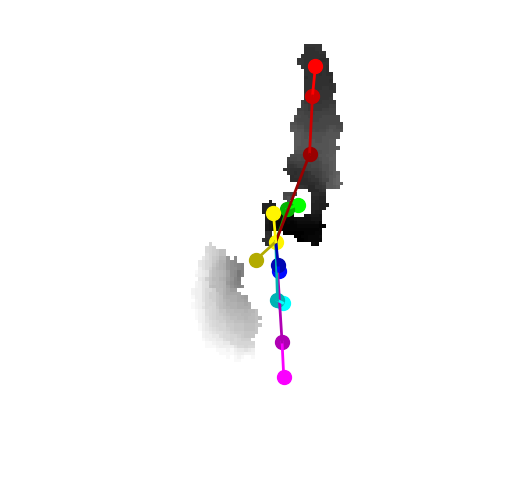}
  \hfill
  \includegraphics[width=0.19\linewidth]{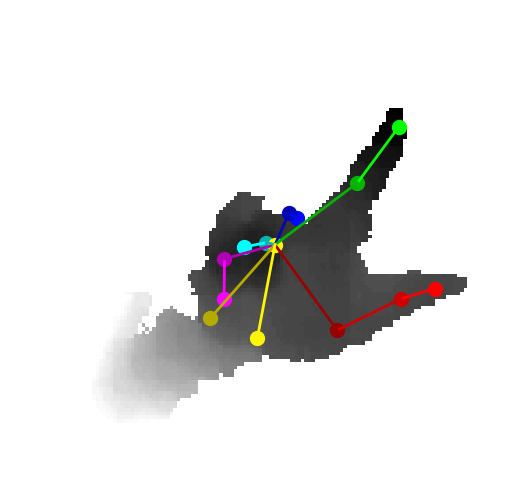}
  \hfill
  \includegraphics[width=0.19\linewidth]{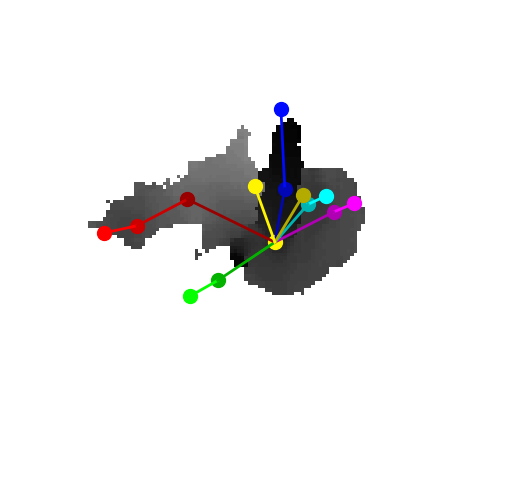}
  \\
  \includegraphics[width=0.19\linewidth]{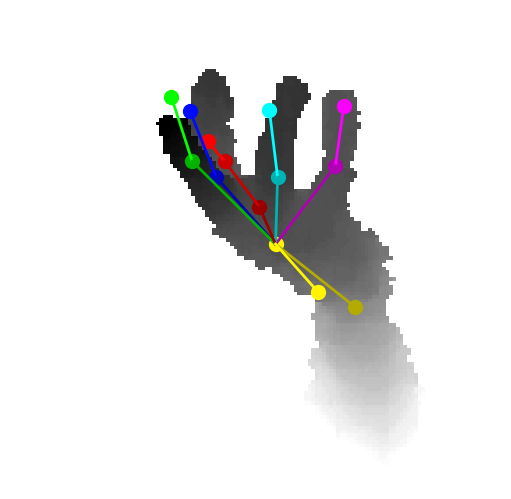}
  \hfill
  \includegraphics[width=0.19\linewidth]{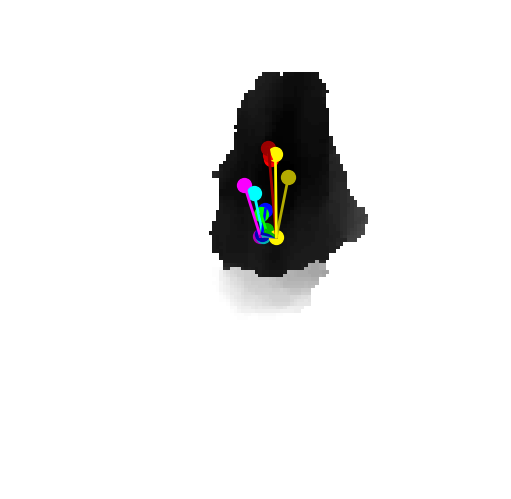}
  \hfill
  \includegraphics[width=0.19\linewidth]{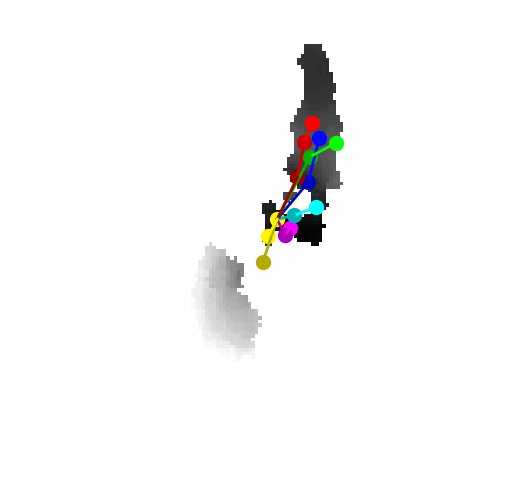}
  \hfill
  \includegraphics[width=0.19\linewidth]{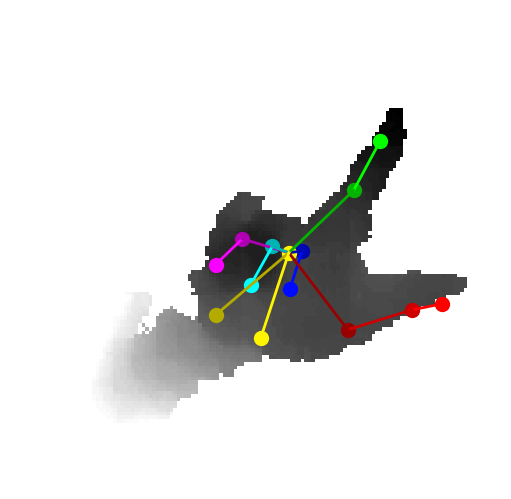}
  \hfill
  \includegraphics[width=0.19\linewidth]{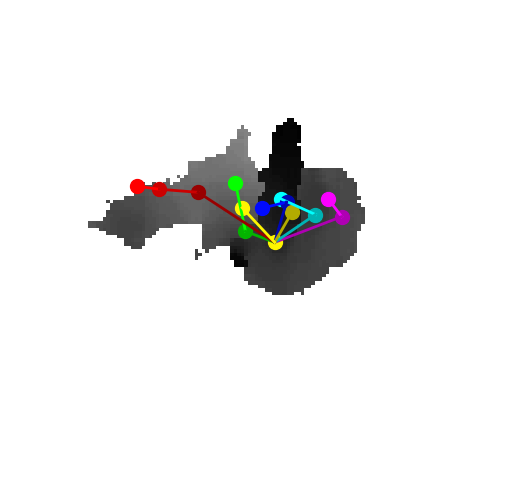}
  \caption[]{\textbf{Frames with largest maximum error.} 
  Test samples overlaid with ground truth (top row) and 
  the predictions of our model (bottom row).
  The errors are mainly due to strongly distorted samples and annotation errors.
  In this case, 79 of the 100 frames with largest maximum error 
  are variations of the three leftmost frames. 
  }
  \label{fig:large_maxerror_frames}
\end{figure*}

\paragraph{Error cases}
We analyze the error cases for our model trained with 100 labeled real samples.
Representative samples from the 100 frames with largest mean error 
are shown in Fig.~\ref{fig:large_meanerror_frames},
samples from the frames with largest maximum error are shown in
Fig.~\ref{fig:large_maxerror_frames}.
We find that our model has problems especially if none of the fingers is clearly 
visible in the depth frame, \ie, the frame has a ''blob-like`` appearance.

\begin{figure*}[t]
  \centering
  \includegraphics[width=0.16\linewidth]{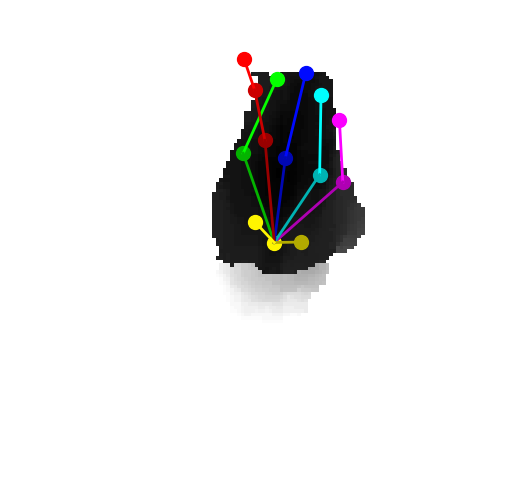}
  \hfill
  \includegraphics[width=0.16\linewidth]{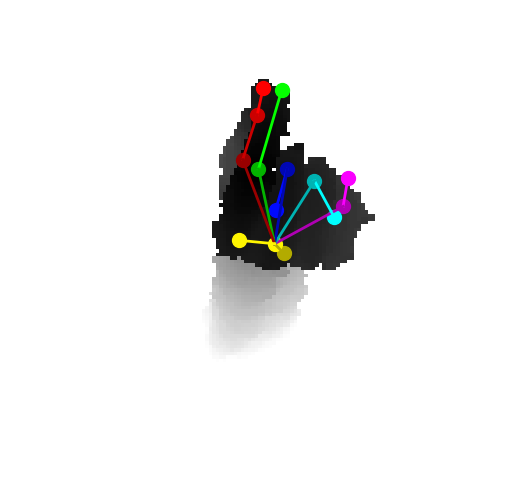}
  \includegraphics[width=0.16\linewidth]{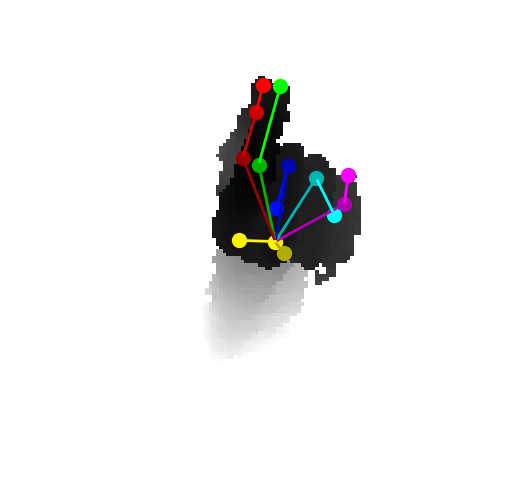}
  \includegraphics[width=0.16\linewidth]{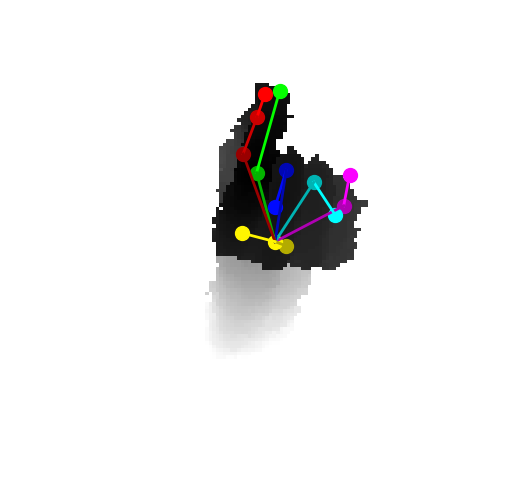}
  \includegraphics[width=0.16\linewidth]{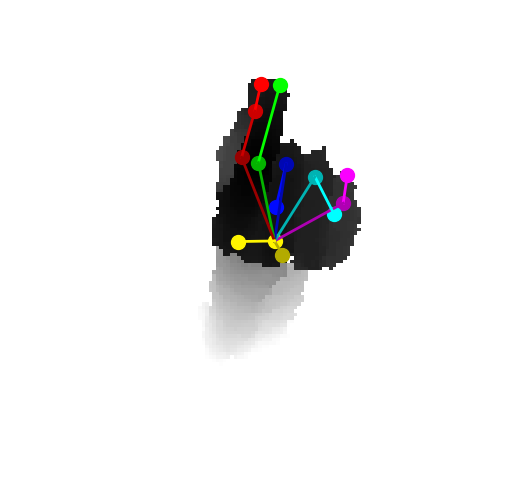}
  \includegraphics[width=0.16\linewidth]{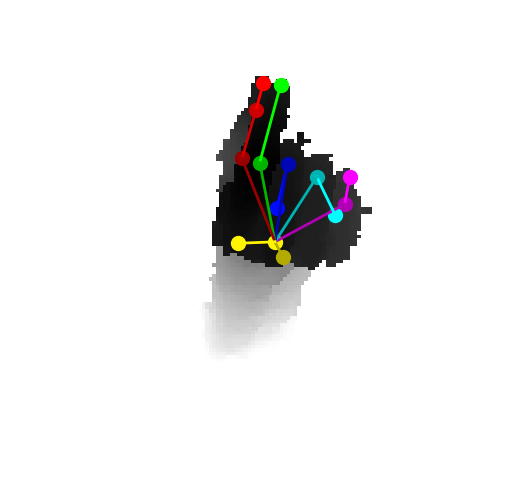}
  \\
  \includegraphics[width=0.16\linewidth]{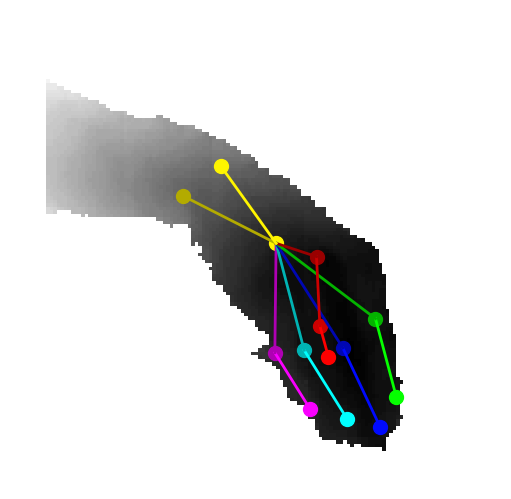}
  \hfill
  \includegraphics[width=0.16\linewidth]{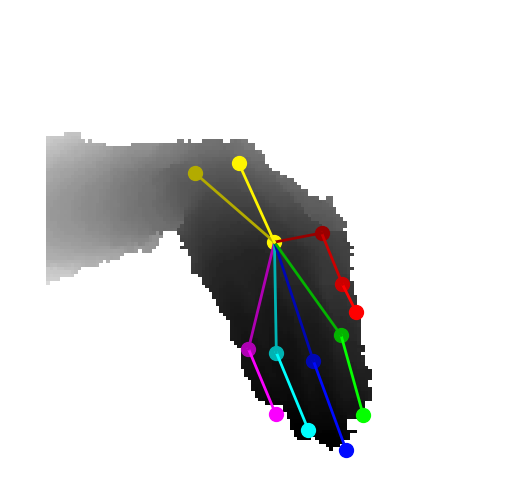}
  \includegraphics[width=0.16\linewidth]{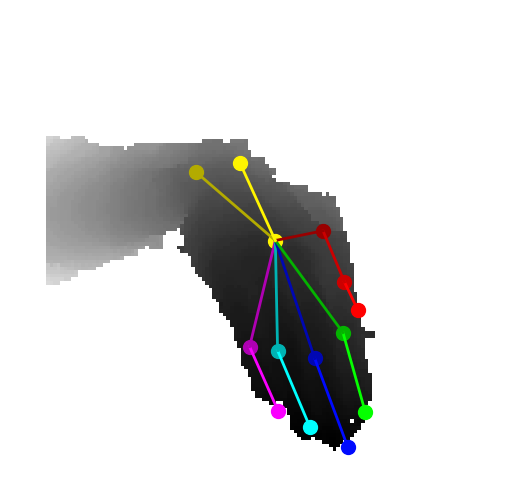}
  \includegraphics[width=0.16\linewidth]{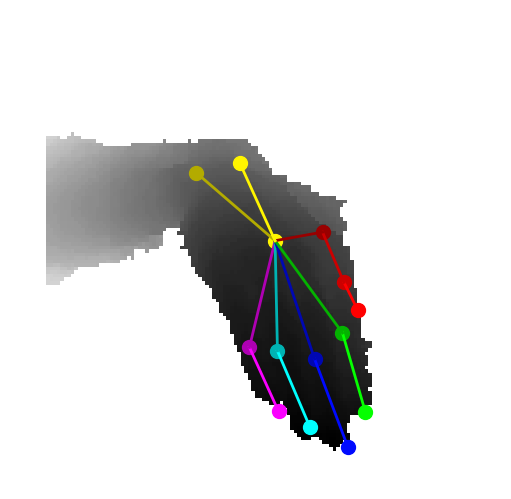}
  \includegraphics[width=0.16\linewidth]{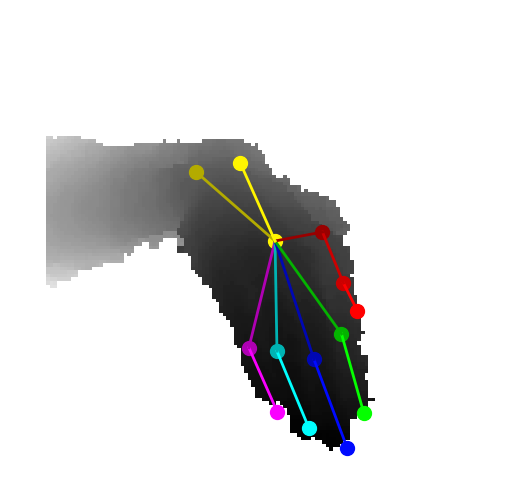}
  \includegraphics[width=0.16\linewidth]{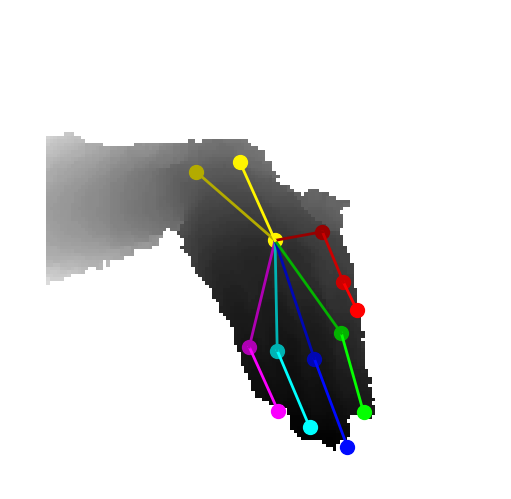}
  \\
  \includegraphics[width=0.16\linewidth]{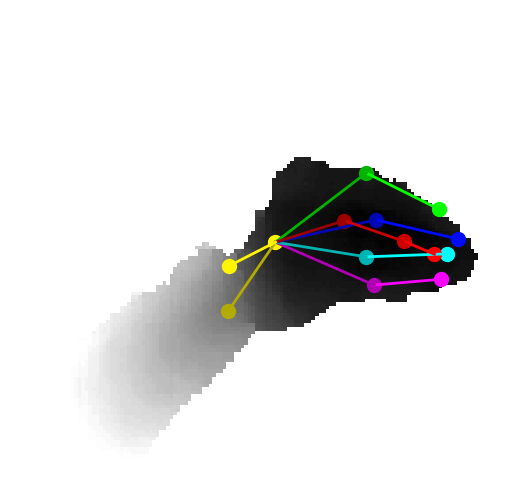}
  \hfill
  \includegraphics[width=0.16\linewidth]{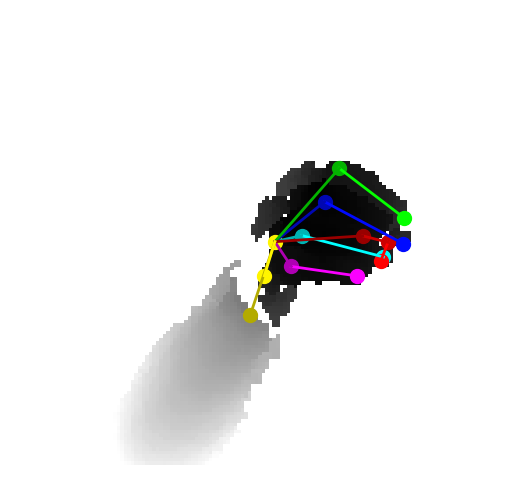}
  \includegraphics[width=0.16\linewidth]{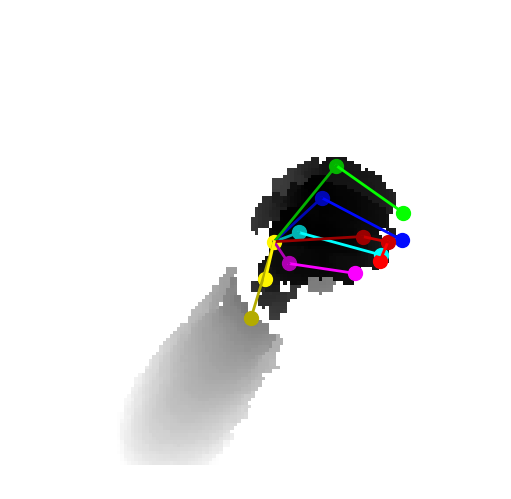}
  \includegraphics[width=0.16\linewidth]{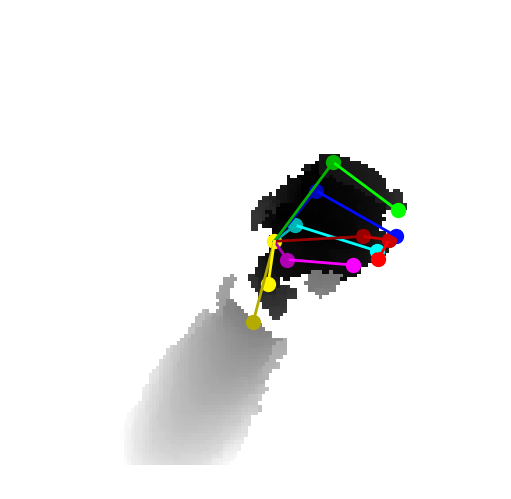}
  \includegraphics[width=0.16\linewidth]{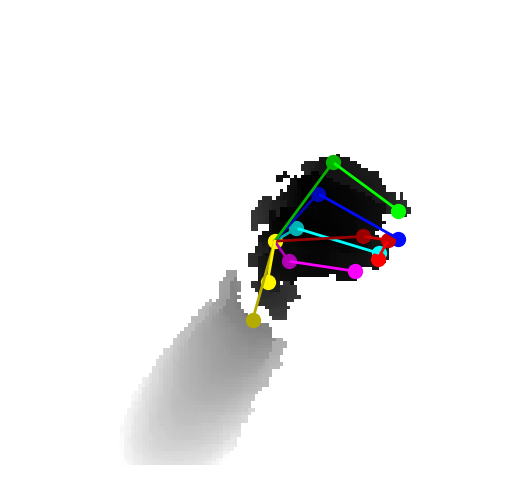}
  \includegraphics[width=0.16\linewidth]{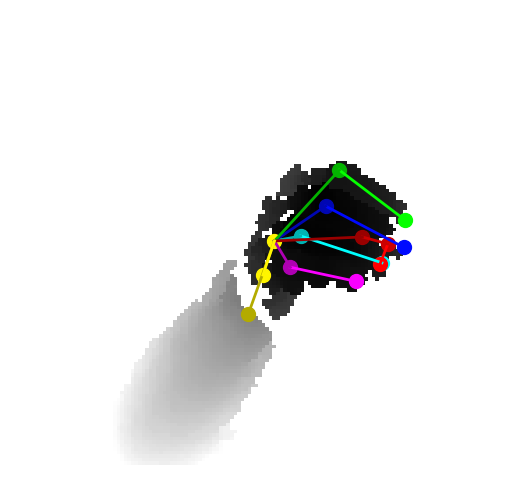}
  \caption[]{\textbf{Nearest neighbors in training set.}
  The test samples with largest error (\cf, Fig.~\ref{fig:large_meanerror_frames} and 
  Fig.~\ref{fig:large_maxerror_frames}) and 
  their nearest neighbors in the training set. 
  Leftmost column shows the test sample, the remaining columns show the 
  corresponding nearest neighbors from the training set.
  Note, the training samples were used unlabeled only.
  }
  \label{fig:nn_in_trainset}
\end{figure*}

For the frames our model had the largest problems with, 
we search for the nearest neighbors in the training set.
We find the nearest neighbors based on the average joint distance 
between the corresponding ground truth annotations 
(after shifting the annotations to the origin).
Fig.~\ref{fig:nn_in_trainset} shows the nearest neighbors for some selected test samples.
We find that for some samples there are no close nearest neighbors in the training set,
and we hypothesize that for such ''blob-like`` structures it is especially difficult 
to obtain valuable feedback from the view prediction objective.
Also note, that the model we are analyzing was trained on only 100 labeled real samples 
and the labels for the nearest neighbors shown in Fig.~\ref{fig:nn_in_trainset} were 
not used.

\end{document}